\documentclass[letterpaper]{article} 
\usepackage{aaai2026}  
\usepackage{times}  
\usepackage{helvet}  
\usepackage{courier}  
\usepackage[hyphens]{url}  
\usepackage{graphicx} 
\usepackage{natbib}  
\usepackage{caption} 
\usepackage{algorithm}
\usepackage{algorithmic}
\usepackage{amsmath}
\usepackage{amssymb}
\usepackage{mathtools}
\usepackage{amsthm}

\theoremstyle{plain}

\newtheorem{proposition}{Proposition}

\theoremstyle{definition}

\theoremstyle{remark}

\usepackage{microtype}
\usepackage{subfigure}
\usepackage{booktabs}
\usepackage{multirow}

\usepackage{lineno}

\usepackage{color}

\usepackage{newfloat}
\usepackage{listings}
\DeclareCaptionStyle{ruled}{labelfont=normalfont,labelsep=colon,strut=off} 
\floatstyle{ruled}
\newfloat{listing}{tb}{lst}{}
\floatname{listing}{Listing}
\title{Diffusion Classifier-Driven Reward for Offline Preference-based \\ 
Reinforcement Learning}

\author {
    Teng Pang\textsuperscript{\rm 1}\equalcontrib,
    Bingzheng Wang\textsuperscript{\rm 1}\equalcontrib,
    Guoqiang Wu\textsuperscript{\rm 1},
    Yilong Yin\textsuperscript{\rm 1},
}
\affiliations {
    \textsuperscript{\rm 1}School of Software, Shandong University, Jinan, Shandong, China\\
    silencept7@gmail.com, binzhwang@gmail.com, guoqiangwu@sdu.edu.cn, ylyin@sdu.edu.cn
}

\usepackage{bibentry}

\begin{document}

\maketitle


\begin{abstract}
Offline preference-based reinforcement learning (PbRL) mitigates the need for reward definition, aligning with human preferences via preference-driven reward feedback without interacting with the environment. 
However, trajectory-wise preference labels are difficult to meet the precise learning of step-wise reward, thereby affecting the performance of downstream algorithms.
To alleviate the insufficient step-wise reward caused by trajectory-wise preferences, we propose a novel preference-based reward acquisition method: Diffusion Preference-based Reward (DPR). 
DPR directly treats step-wise preference-based reward acquisition as a binary classification and utilizes the robustness of diffusion classifiers to infer step-wise rewards discriminatively. 
In addition, to further utilize trajectory-wise preference information, we propose Conditional Diffusion Preference-based Reward (C-DPR), which conditions on trajectory-wise preference labels to enhance reward inference.
We apply the above methods to existing offline RL algorithms, and a series of experimental results demonstrate that the diffusion classifier-driven reward outperforms the previous reward acquisition method with the Bradley-Terry model.
\end{abstract}


\section{Introduction}
\label{introduction}

Reinforcement learning (RL)~\cite{sutton2018reinforcement}, as a pivotal field within machine learning, fuels advances in diverse domains and real-world applications such as robot control~\cite{majumdar2023we, black2024pi_0}, autonomous driving~\cite{elallid2022comprehensive, huang2022efficient}, large language models~\cite{NEURIPS2022_b1efde53, rafailov2024direct}, and so on. The core content of RL is to continuously improve the decision-making ability of agents through the use of reward feedback mechanisms, thereby achieving goals in specific scenarios~\cite{sutton2018reinforcement, franccois2018introduction,pmlr-v80-haarnoja18b}. However, due to the complexity of the real world, defining a special reward mechanism often poses a significant challenge. To alleviate this issue of reward specification, the preference-based reinforcement learning (PbRL) paradigm is proposed~\cite{christiano2017deep, wirth2017survey}. 

In PbRL, agents are equipped with paired trajectory preference data, enabling them to learn an informative reward function that serves as feedback. 
The reward functions are often learned with the Bradley-Terry model~\cite{bradley1952rank} and seamlessly integrated into standard RL algorithms, which 
are subsequently applied to specific tasks.
This framework has been fully utilized in Large Language Models~\cite{NEURIPS2022_b1efde53, dai2024safe}.
However, 
frequent online feedback collection
has resulted in a significant and unwarranted drain on human resources, especially in some complex and challenging interaction scenarios.
Therefore, research on PbRL in offline scenarios has gradually gained attention~\cite{shin2021offline}.

For offline PbRL, existing approaches continue to employ the Bradley-Terry model for establishing human preference 
(e.g., MLP-based Markovian rewards~\cite{christiano2017deep, wirth2017survey} or Transformer-based non-Markovian rewards~\cite{kim2023preference}) and minimizing the discrepancy between trajectory-wise human preference and model predictions to learn the reward function~\cite{tu2024dataset}. 
Although the trajectory-wise preferences modeled in the above manner
may partially fulfill human preference requirements, they still face certain challenges in learning an effective step-wise reward~\cite{zhang2023flow}.
For instance, a clearly defined step-wise reward can imply trajectory-wise preferences; however, trajectory-wise preferences might not yield a unique step-wise reward.
Recently, some work has eliminated the need for reward functions by directly optimizing downstream policy using preference information~\cite{hejna2024inverse}. 
Deviating from the traditional Markov Decision Process (MDP) with a fixed reward function, these methods may exhibit instability during training and are more prone to generating out-of-distribution (OOD) data~\cite{pmlr-v235-xu24h}. 

To address the above issues, we propose a reward acquisition method based on the diffusion model, named Diffusion Preference-based Reward (\textbf{DPR}). 
Through the diffusion classifier, DPR infers step-wise rewards by distinguishing data under two types of step-wise label supervision, enabling effective downstream policy learning.
However, rigidly separating step-wise reward may be too absolute, potentially missing valuable reward information conveyed by the preference labels.
Based on it, we incorporate preference labels into the classification, further proposing Conditional Diffusion Preference-based Reward (\textbf{C-DPR}) to mitigate the impact of step-wise absolute classification. 

To demonstrate the effectiveness of preference reward with the diffusion model, we apply our methods to existing offline RL algorithms and evaluate them on pre-collected preference datasets. 
The results of the experiments show that, compared to using the Bradley-Terry model(e.g., MLP-based Markovian rewards or Transformer-based non-Markovian rewards), 
our methods 
exhibits significant performance advantages. 

In summary, our contributions are as follows: 
\begin{itemize}
    \item We propose Diffusion Preference-based Reward (DPR), a diffusion classifier-based method for inferring step-wise rewards to support offline RL.
    \item To further utilize trajectory-wise preference information, we propose Conditional Diffusion Preference-based Reward (C-DPR), which infers step-wise rewards conditioned on (trajectory-wise) preference labels.
    \item We apply DPR and C-DPR to various algorithms in offline RL, and experiments show that they outperform those obtained from the use of the Bradley-Terry model.
\end{itemize}

\section{Preliminaries}
\label{setting}

\noindent \textbf{Offline Reinforcement Learning.} \
Offline reinforcement learning (RL), also known as batch RL, mainly solves how to improve an agent without interacting with the environment. In this setting, we consider an infinite-horizon Markov Decision Process (MDP) as a tuple ($\mathcal{S}$, $\mathcal{A}$, $\mathcal{T}$, $r$, $\rho_0$, $\gamma$). Here, $s\in\mathcal{S}$ and $a\in\mathcal{A}$ represent the state and action, $\mathcal{T}:\mathcal{S} \times \mathcal{A} \times \mathcal{S} \to [0,1]$ and $r: \mathcal{S} \times \mathcal{A} \to \mathbb{R}$ are the transition function and (step-wise) reward function, and the initial state distribution $\rho_0$ and discount factor $\gamma \in (0,1]$ control the environment's starting state and reward decay, respectively.
Unlike online RL interacting with the environment, we are given a dataset $\mathcal{D}_{\mathrm{off}} = \left \{ (s_{i}, a_{i}, r_i,s'_{i}) \right \}_{i=1}^{I}$ with $I$ tuples, where these data are obtained by sampling from one or more behavior policies. 
Under these conditions, we aim to learn a policy $\pi:\mathcal{S} \to \Delta(\mathcal{A})$ to maximize the discounted accumulated rewards $J(\pi):=\mathbb{E}_{s_0\sim \rho_0, \pi}[\sum_{h=0}^{\infty}\gamma^tr(s_h,a_h)]$.

\noindent \textbf{Offline Preference-based Reinforcement Learning.}
Based on offline RL, offline Preference-based Reinforcement Learning (PbRL) introduces labelled data with human preferences, hoping to learn policy from human preferences~\cite{shin2021offline, kim2023preference}. 
Generally speaking, considering each trajectory $\tau = \left \{ (s_h, a_h)\right \}_{h=0}^{H-1}$ of length $H$, 
offline PbRL typically includes two types of datasets: 
$\mathcal{D}^{\mathrm{L}}=\left \{ (\tau^{(0)}_{n}, \tau^{(1)}_{n}, y_{n}) \right \}_{n=1}^{N}$ with a few trajectory-wise labels of human preference, and $\mathcal{D}^{\mathrm{U}}=\left \{ \tau_{m} \right \}_{m=1}^{M}$ without any labels. For $\mathcal{D}^{\mathrm{L}}$, we set $y_{n}=1$ if $\tau^{(1)}_{n} \succ \tau^{(0)}_{n}$, $y_{n}=0$ if $\tau^{(0)}_{n} \succ \tau^{(1)}_{n}$, and $y_{n}=0.5$ otherwise\footnote{We set $\tau^a \succ \tau^b$ to indicate that trajectory $\tau^a$ is preferable to trajectory $\tau^b$.}. 

Due to rewards are not explicitly provided in offline PbRL,
preference data are often used to train a step-wise reward model $r_{\phi}$, which subsequently serves downstream offline RL algorithms.
In order to learn $r_{\phi}$, previous work often construct human preference based on the Bradley-Terry model~\cite{bradley1952rank}:
\begin{align}\label{define:BTmodel}
    P[\tau^{(i)} & \succ\tau^{(j)}] = \notag \\
      &\frac{\exp\sum_{h} r_{\phi}(s^{(i)}_h,a^{(i)}_h)}{\exp\sum_{h} r_{\phi}(s^{(i)}_h,a^{(i)}_h) + \exp\sum_{h} r_{\phi}(s^{(j)}_h,a^{(j)}_h)}.
\end{align}
Then, to align the above distribution with preference labels $y$, the reward model is typically learned by minimizing binary cross-entropy loss:
\begin{align}\label{define:reward learning}
    \mathcal{L}&_{\mathrm{BT}}:=-  \mathbb{E}_{(\tau^{(0)}, \tau^{(1)}, y)\sim \mathcal{D}^L} \notag \\ 
    & \Big[  (1-y)\log P[\tau^{(0)}\succ\tau^{(1)}]  + y\log P[\tau^{(1)}\succ\tau^{(0)}] \Big].
\end{align}
For simplicity, we assume the preference of $\tau^{(1)}_{n}$ is not weaker than the preference of $\tau^{(0)}_{n}$ in this work, i.e., $\tau^{(1)}_{n} \succeq \tau^{(0)}_{n}$ and set $\tilde{\mathcal{D}}^{\mathrm{L}}=\left \{ (\tau^{(0)}_{n}, \tau^{(1)}_{n}) \right \}_{n=1}^{N}$. 

\noindent \textbf{Denoising Diffusion Probability Model.}
The diffusion model~\cite{ho2020denoising,song2020score}, which is a powerful generative model, 
involves initially converting the target data $x_0$ sampled from a dataset $\mathcal{D}$ into random noise $\epsilon\sim \mathcal{N}(0,\textit{\textbf{I}})$ 
through the forward noising process, followed by a reverse denoising process to progressively sample and restore the target distribution. 
For diffusion models with discrete timesteps, we generate noise based on gradually restored noisy data 
and train the network with the evidence lower bound (ELBO) of target distribution $p(x_0)$:
\begin{align}\label{define:diffusion}
    \mathcal{L}_{\mathrm{DM}} (\phi) & = \mathbb{E}_{x_0\sim \mathcal{D},t\sim U(1, T), \epsilon\sim \mathcal{N}(0,\textit{\textbf{I}})}\notag \\
    &[||\epsilon-\epsilon_{\phi}(\sqrt{\bar{\alpha}_t}x_0+\sqrt{1-\bar{\alpha}_t}\epsilon, t)||^2],
\end{align}
where $T$ is the total diffusion timestep, $U$ denotes the uniform distribution,
$\epsilon_{\phi}$ denotes the diffusion network with the parameter $\phi$, and $\bar{\alpha}_t$ is a predefined noise schedule related to $t$, respectively. For simplicity, we set diffusion model $\epsilon_{\phi}(x_0,\epsilon,t)=\epsilon_{\phi}(\sqrt{\bar{\alpha}_t}x_0+\sqrt{1-\bar{\alpha}_t}\epsilon, t)$. 


\noindent \textbf{Diffusion Classifiers.}
The Diffusion Classifier utilizes pre-trained conditional diffusion models $\epsilon_{\phi}(x_0,\epsilon,t; c)$, conditioned on $c\in C$, to enable robust classification~\cite{10376944, NEURIPS2024_59a3444d, pmlr-v235-chen24k}.
It uses the ELBO to approximate the conditional log-likelihood $\log p(x_0|c)$ and applies Bayes' theorem to compute the classification probability:
\begin{align}\label{define: diffusion classifier}
    p(c|x_0) \approx \frac{\exp\{-\mathbb{E}_{\epsilon,t}[||\epsilon-\epsilon_{\phi}(x_0,\epsilon,t;c)||^2]\}}{\sum_{\hat{c}\in C}\exp\{-\mathbb{E}_{\epsilon,t}[||\epsilon-\epsilon_{\phi}(x_0,\epsilon,t;\hat{c})||^2]\}}.
\end{align}

\begin{figure*}
    \centering
    \includegraphics[width=0.9\linewidth]{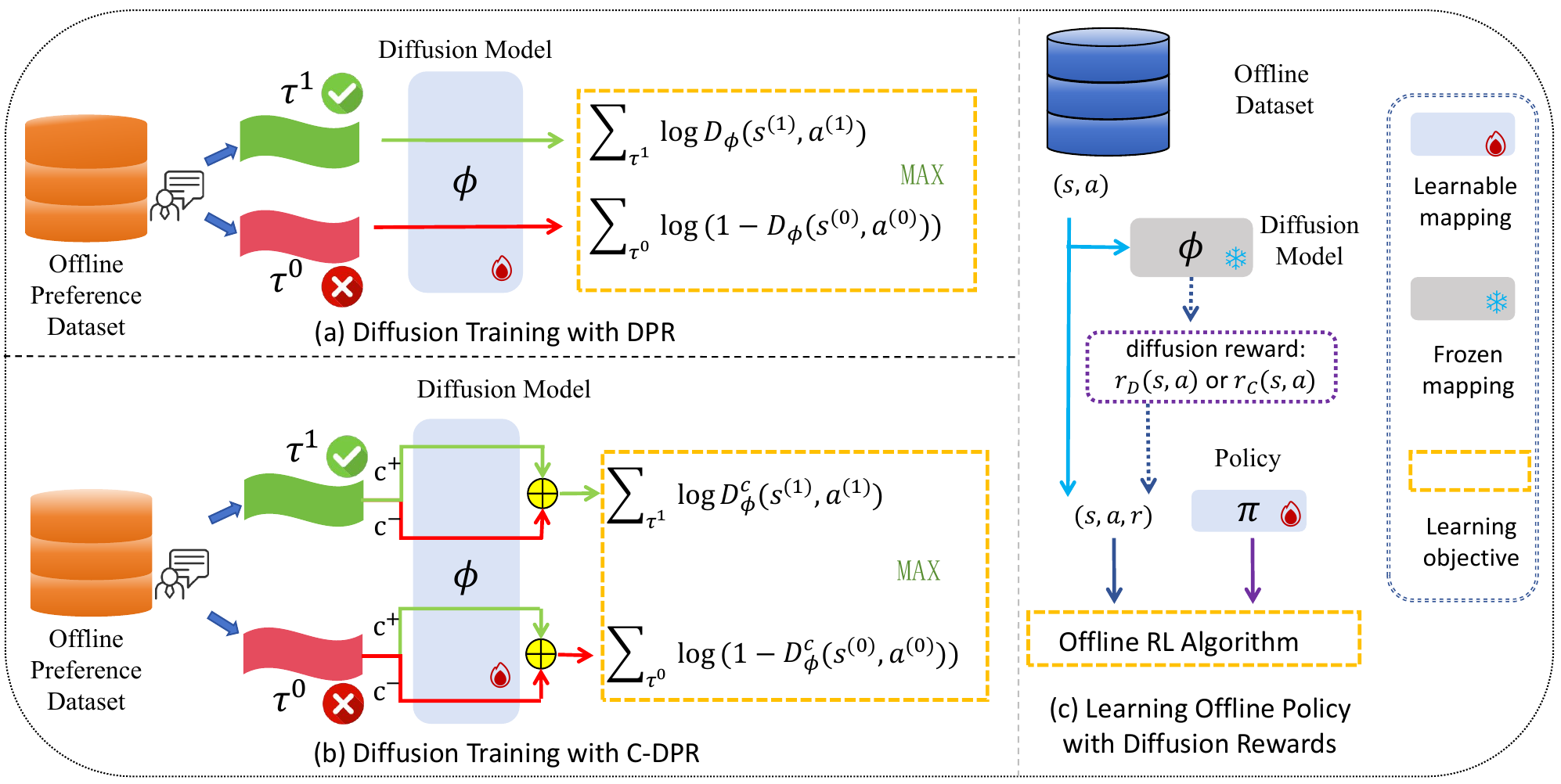}
    \caption{Overview of DPR and C-DPR training and inference. 
    (a) DPR: Directly using the diffusion model to classify $(s,a)$ with its associated preference label.
    (b) C-DPR: Using the conditional diffusion model to classify $(s,a)$ with all labels.
    (c) Offline RL with diffusion reward.
    }
    \label{fig:DPR}
\end{figure*}

\begin{algorithm}[tb]
   \caption{Offline PbRL with Diffusion Reward}
   \label{alg:DPR}
   \textbf{Input}: preference dataset $\tilde{\mathcal{D}}^\mathrm{L}$, unlabeled dataset $\mathcal{D}^\mathrm{U}$; diffusion model $\epsilon_{\phi}$, offline policy $\pi_{\theta}$; diffusion training epoch $K$, diffusion timestep $T$ \\
   \textbf{Output}: offline policy $\pi_{\theta}$
   \begin{algorithmic}[1]
   \STATE Initialize $\mathcal{D}_{\mathrm{off}} \gets  \mathcal{D}^\mathrm{U}$
   \STATE Method = `DPR' or `C-DPR'
   \FOR{$k=0$ {\bfseries to} $K-1$}
   \STATE Sample $(\tau^{(0)}, \tau^{(1)})\sim \tilde{\mathcal{D}}^\mathrm{L}$
   \STATE Sample $t\sim U(1, T),\epsilon\sim\mathcal{N}(0,\textit{\textbf{I}})$
   \IF {Method = `DPR'} 
   \STATE Update diffusion $\epsilon_{\phi}$ by Eq.~\eqref{DPR} // DPR
   \ELSE 
   \STATE Update diffusion $\epsilon_{\phi}$ by Eq.~\eqref{C-DPR} // C-DPR
   \ENDIF 
   \ENDFOR
   \IF {Method = `DPR'} 
   \STATE Compute reward $r=r_{\mathrm{D}}(s,a)$ by Eq.~\eqref{reward:DPR} with $\mathcal{D}_{\mathrm{off}}$
   \ELSE 
   \STATE Compute reward $r=r_{\mathrm{C}}(s,a)$ by Eq.~\eqref{reward:C-DPR} with $\mathcal{D}_{\mathrm{off}}$
   \ENDIF
   \STATE $\mathcal{D}_{\mathrm{off}} \gets (\mathcal{D}_{\mathrm{off}},\{r\})$
   \STATE Update offline policy $\pi_{\theta}$ with $\mathcal{D}_{\mathrm{off}}$ and offline RL algorithm (e.g., CQL, IQL, TD3BC)
\end{algorithmic}
\end{algorithm}

\section{Methodology}
\label{methodology}

In this section, we introduce the preference-based step-wise reward acquisition with diffusion models. 
We assume that step-wise data come from two reward-based distributions and treat reward acquisition as binary classification.
Based on this, we propose two diffusion-based reward acquisition methods, DPR and C-DPR, designed for different preference reward modeling. Subsequently, we infer step-wise rewards via the ELBO of diffusion and apply them to offline RL algorithms. 
We present the overall framework in Figure~\ref{fig:DPR}.

\subsection{Diffusion Preference-based Reward}
\label{Section:Joint Reward}

In offline PbRL, reward learning relies heavily on preference labels. Due to the localized nature of preference information, modeling methods based on (MLP-based) Markovian reward~\cite{christiano2017deep, wirth2017survey} and (Transformer-based) non-Markovian reward~\cite{kim2023preference} often infer preference distributions through the Bradley-Terry model, followed by binary classification optimization~\cite{tu2024dataset}. 
However, as preference labels are provided at the trajectory level, directly leveraging them to learn step-wise reward functions can introduce non-negligible bias. 

To address the above issues, we need to extract step-wise rewards from the trajectory-wise preference label. 
We first assume that the step-wise data $ (s,a)$ comes from two different distributions: a high-reward distribution $\mathcal{D}_{\mathrm{high}}$, labeled as $y=1$, and a low-reward distribution $\mathcal{D}_{\mathrm{low}}$, labeled as $y=0$. 
Based on this assumption, we can train a step-wise classifier $D$ using binary cross-entropy to predict $p(y=1|(s,a))$.
\begin{align}
    \max_{D} \ & \mathbb{E}_{(\hat{s}^h,\hat{a}^h) \sim \mathcal{D}_{\mathrm{high}}} \Big[ \log  D(\hat{s}^h,\hat{a}^h) \Big] \notag \\
    \label{Croee-Entropy}
   &  + \mathbb{E}_{(\hat{s}^l,\hat{a}^l) \sim \mathcal{D}_{\mathrm{low}}} \Big[\log \Big(1-D(\hat{s}^l,\hat{a}^l)\Big)\Big].
\end{align}
In offline PbRL, where rewards are derived from preference dataset $\tilde{\mathcal{D}}^{\mathrm{L}}$, we regard high- and low-preference data from $\mathcal{D}_{\mathrm{high}}$ and $\mathcal{D}_{\mathrm{low}}$, respectively. This allows us to train the classifier $D$ to predict $p(y=1|(s,a))$ as follows:
\begin{align}
    \max_D \ & \mathbb{E}_{(\tau^{(0)},\tau^{(1)})\sim \tilde{\mathcal{D}}^{\mathrm{L}}}   \Bigg [  \frac{1}{H} \sum_{(s^{(1)}_h, a^{(1)}_h)\in \tau^{(1)}}  \log \Big( D(s^{(1)}_h, a^{(1)}_h) \Big) \notag \\
    \label{DPR_classifier}
    & +  \frac{1}{H} \sum_{(s^{(0)}_h, a^{(0)}_h)\in \tau^{(0)}} \log \Big(1-D (s^{(0)}_h,a^{(0)}_h) \Big) \Bigg ] .
\end{align}

However, since trajectory-wise preference labels are inherently uncertain, step-wise binary classification from the preference dataset is challenging.
Inspired by the robust classification capabilities of diffusion classifiers~\cite{10376944,  NEURIPS2024_59a3444d, pmlr-v235-chen24k}, we propose Diffusion Preference-based Reward (\textbf{DPR}) for step-wise reward acquisition, aiming to alleviate the uncertainty caused through trajectory-wise preference labelling.
In addition, by define $q(s,a) = p((s,a)|y=1)$, we find $p(y=1|(s,a)) \propto q(s,a) \approx \exp\{-\mathbb{E}_{\epsilon,t}[||\epsilon-\epsilon_{\phi}((s,a),\epsilon,t)||^2]$ through the Bayes' theorem
\footnote{Considering the diffusion model with its ELBO defined in Eq.~\eqref{define:diffusion}~\cite{ho2020denoising, 10376944}, we have $q(s, a) \approx \exp\{-\mathbb{E}_{\epsilon,t}[||\epsilon-\epsilon_{\phi}((s,a),\epsilon,t)||^2]\}$}.  
We can therefore train the diffusion model without conditioning, using the above approximation for prediction. 
Formally, based on the the preference trajectory dataset $\tilde{\mathcal{D}}^{\mathrm{L}}$, we train the diffusion model $D_{\phi}$ by:
\begin{align}
    \max_{\phi} \ & \mathbb{E}_{(\tau^{(0)},\tau^{(1)})\sim \tilde{\mathcal{D}}^{\mathrm{L}}}   \Bigg [ \frac{1}{H} \sum_{(s^{(1)}_h, a^{(1)}_h)\in \tau^{(1)}} \log \Big( D_{\phi}(s^{(1)}_h, a^{(1)}_h) \Big) \notag \\
    \label{DPR}
    &+ \frac{1}{H} \sum_{(s^{(0)}_h, a^{(0)}_h)\in \tau^{(0)}} \log \Big(1-D_{\phi}(s^{(0)}_h,a^{(0)}_h) \Big) \Bigg ],
\end{align}
where 
$D_{\phi}(s, a)=\exp\{-\mathbb{E}_{\epsilon,t}[||\epsilon-\epsilon_{\phi}((s,a),\epsilon,t)||^2]\}$. 

Furthermore, by optimizing Eq.\eqref{DPR}, we can also obtain a good trajectory-wise binary classifier, just like the Bradley-Terry model,
which is advantageous for capturing human preferences.
Specifically, considering $\tau = \left \{ (s_h, a_h)\right \}_{h=0}^{H-1}$, 
we can set the average of step-wise predictions as the overall trajectory prediction with trajectory-wise classifier $D_{\tau}$, i.e., $D_{\tau}(\tau) = \frac{1}{H} \sum_{(s_h, a_h)\in \tau} D_{\phi}(s_h, a_h)$. 
To bridge the optimization between trajectory-wise and step-wise classifiers, we propose the following proposition based on the concavity of the log function and Jensen's inequality:
\begin{proposition}\label{proposition:logarithmic}
Given a set of variables $x_0,x_1,...,x_{H-1}$ with length $H $, where $x_{h}\in (0,1)$ for \ $\forall h \in[0,H-1]$, the following hold:
\begin{align}
    \log \left( \frac{1}{H} \sum_{h=0}^{H-1} x_h \right) \geq \frac{1}{H}\sum_{h=0}^{H-1}\log(x_h).
\end{align}
\end{proposition}

According to the Proposition~\ref{proposition:logarithmic}, we can show that optimizing Eq.\eqref{DPR} effectively optimizes its upper bound, thereby achieving binary classification over trajectories:
\begin{align}
    \label{DPR_trajectory}
    & \max_{\phi} \mathbb{E}_{(\tau^{(0)}, \tau^{(1)})\sim \tilde{\mathcal{D}}^{\mathrm{L}}}  
    \Big[ \log \Big( \frac{1}{H} \sum_{(s^{(1)}_h, a^{(1)}_h)\in \tau^{(1)}} D_{\phi}(s^{(1)}_h, a^{(1)}_h) \Big) \notag \\
    & \quad \ + \log \Big( \frac{1}{H} \sum_{(s^{(0)}_h, a^{(0)}_h)\in \tau^{(0)}} \Big( 1-D_{\phi}(s^{(0)}_h,a^{(0)}_h) \Big) \Big) \Big], \\
    & \Rightarrow 
    \label{classifier_trajectory}
    \max_{D_{\tau}} \ \mathbb{E}_{(\tau^{(0)}, \tau^{(1)}) \sim \tilde{\mathcal{D}}^{\mathrm{L}}} \Big[ \log  D_{\tau}(\tau^{(1)}) 
    + \log (1-D_{\tau}(\tau^{(0)}))\Big].
\end{align}

As shown above, when we perform classification using DPR (i.e., Eq.\eqref{DPR}), it also implies that preference trajectories can be properly classified, thereby enabling the step-wise rewards to align with trajectory-wise labels.
Besides, the optimization loss is directly applied to step-wise reward estimation, thereby improving the stability of the obtained reward.
We describe the training process of DPR in Figure~\ref{fig:DPR} (a).

\subsection{Conditional DPR}\label{Section:Conditional Reward}
While DPR classifies preference trajectories using step-wise binary labels, it does not consider all labels for a single pair $(s,a)$ like diffusion classifiers (e.g., Eq.~\eqref{define: diffusion classifier}) and thus may fail to discriminate them fully. 
For example, when evaluating all preference trajectories collectively, certain high-preference trajectories may appear less favourable than some low-preference ones. 

To further utilize trajectory-wise preference labels for effective classification, following Eq.~\eqref{define: diffusion classifier}, we propose Conditional Diffusion Preference-based Reward (\textbf{C-DPR}), which predicts $p(y=1|(s,a))$ with the conditional diffusion:
\begin{align}
    \max_{\phi} \ & \mathbb{E}_{(\tau^{(0)}, \tau^{(1)})\sim \tilde{\mathcal{D}}^{L}}   \Bigg[ \frac{1}{H} \sum_{(s^{(1)}_h, a^{(1)}_h)\in \tau^{(1)}} \log \Big( D^c_{\phi}(s^{(1)}_h, a^{(1)}_h) \Big) \notag \\
    \label{C-DPR}
    & + \frac{1}{H} \sum_{(s^{(0)}_h, a^{(0)}_h)\in \tau^{(0)}} \log \Big( 1-D^c_{\phi}(s^{(0)}_h,a^{(0)}_h) \Big) \Bigg].
\end{align}
where $D^c_{\phi}(s, a)=\frac{D^{c}(s,a;\mathbf{c^{+}})} {D^{c}(s,a;\mathbf{c^{+}}) + D^{c}(s,a;\mathbf{c^{-}})}$ and $D^{c}(s,a;\mathbf{c})=\exp\{-\mathbb{E}_{\epsilon,t}[||\epsilon-\epsilon_{\phi}((s,a),\epsilon,t;\mathbf{c})||^2]\}$. 
Here, $\mathbf{c}$ signifies preference label, with $\mathbf{c^+}=1$ indicating the higher preference and $\mathbf{c^-}=0$ indicating the lower one. 

In contrast to Eq.\eqref{DPR}, Eq.\eqref{C-DPR} conditions on preference labels to construct a binary diffusion classifier for obtaining step-wise reward.
Note that, unlike modeling trajectory-wise preferences using the Bradley-Terry model, Eq.~\eqref{C-DPR} builds a step-wise one by inferring rewards for individual state-action pairs across preferences. 
We describe the training process of C-DPR in Figure~\ref{fig:DPR} (b).

\subsection{Offline RL with Diffusion Reward}
\label{Section:Reward Application}

For downstream offline RL algorithms, we assign rewards to each step-wise pair in the unlabeled offline dataset based on the learned diffusion model.
Following GAIL~\cite{NIPS2016_cc7e2b87}, which uses discrimination results as rewards for reinforcement learning, we use the average exponential form of the ELBO over all time steps as the reward, similar to DiffAIL~\cite{wang2024diffail}. It is computed as follows:
\begin{align}
   \label{reward:DPR} & \mathrm{\textbf{DPR}}: r_{\mathrm{D}}(s,a) = -\frac{1}{T}\sum_{t=1}^{T} \log \Big( 1 - \bar{D}_{\phi}(s,a,t) \Big), \\ 
   \label{reward:C-DPR} & \mathrm{\textbf{C-DPR}}: r_{\mathrm{C}}(s,a) = -\frac{1}{T}\sum_{t=1}^{T} \log \Big( 1 - \bar{D}_{\phi}^c(s,a,t) \Big),
\end{align}
where $\bar{D}_{\phi}(s,a,t)=\exp\{-\mathbb{E}_{\epsilon}[||\epsilon-\epsilon_{\phi}((s,a),\epsilon,t)||^2]\}$, $\bar{D}^c_{\phi}(s, a, t)=\frac{\bar{D}^{c}(s,a,t;\mathbf{c^{+}})} {\bar{D}^{c}(s,a,t;\mathbf{c^{+}}) + \bar{D}^{c}(s,a,t;\mathbf{c^{-}})}$ and $\bar{D}^{c}(s,a,t;\mathbf{c})=\exp\{-\mathbb{E}_{\epsilon}[||\epsilon-\epsilon_{\phi}((s,a),\epsilon,t;\mathbf{c})||^2]\}$.
After obtaining the reward, we can leverage existing offline RL algorithms to learn the policy $\pi$, as shown in Figure~\ref{fig:DPR} (c).
Algorithm~\ref{alg:DPR} demonstrates the entire process.

\begin{table*}[t]
\centering
\small
\setlength{\tabcolsep}{4pt}{
\begin{tabular}{c|cccccc|cccccc}
\toprule 
\multirow{2}{*}{Dataset}&\multicolumn{6}{|c|}{IQL}&\multicolumn{6}{|c}{TD3BC}\\
\cmidrule{2-13}
& MLP& MLP*& TFM& TFM*& DPR& C-DPR
& MLP& MLP*& TFM& TFM*& DPR& C-DPR\\
\midrule  
HalfCheetah-M&
43.01& 43.3&  44.24& 43.24& {\color{blue}44.90} &  \textbf{\color{blue}45.57} &
42.31&  34.8& \textbf{46.91}$^{\dagger}$& \textbf{46.62}$^{\dagger}$ & 44.10& 45.09\\
HalfCheetah-MR&  
36.90& 38.0&  40.83& 39.49& {\color{blue}42.79} & \textbf{\color{blue}44.09}$^{\dagger}$ &
39.65&  38.9& 41.71& 29.58& 40.38& \textbf{\color{blue}43.48}$^{\dagger}$\\
HalfCheetah-ME&  
91.48& 91.0&  91.76& 92.20& 87.99& \textbf{\color{blue}93.62} &
70.69&  73.8& \textbf{93.49}& 80.83& \textbf{93.86} & \textbf{\color{blue}94.58}$^{\dagger}$\\
\midrule
Hopper-M&    
53.82& 50.8&  62.78& 67.81& \textbf{\color{blue}74.49} & {\color{blue}70.98} &
10.00&  48.0& 22.35& \textbf{99.42}$^{\dagger}$ & 83.54& 90.99\\
Hopper-MR&   
91.74& 87.1&  \textbf{88.77} & 22.65& 87.53& 86.10&
46.68&  25.8& 87.07& 41.44& \textbf{\color{blue}92.62}$^{\dagger}$ & {\color{blue}90.72}\\
Hopper-ME&    
 81.73& 94.3&  105.10& \textbf{111.43}$^{\dagger}$& 108.44& 99.34&
65.52&  97.4& 29.47&  91.18& \textbf{\color{blue}101.95}& 96.56\\
\midrule 
Walker2D-M&    
74.12& 78.4&  81.57& 79.36& {\color{blue}83.47} & \textbf{\color{blue}85.77}$^{\dagger}$ &
46.00&  26.3& 59.41& 84.11& 82.43& \textbf{\color{blue}85.70}$^{\dagger}$\\
Walker2D-MR&  
69.58& 67.3&  56.92& 56.52& {\color{blue}67.23} & \textbf{\color{blue}75.95}$^{\dagger}$ &
21.21&  47.2& 33.78& 61.94& {\color{blue}67.10} & \textbf{\color{blue}73.69}\\
Walker2D-ME&   
109.02& 109.4&  109.07& 109.12& \textbf{\color{blue}111.89}$^{\dagger}$& \textbf{\color{blue}111.98}$^{\dagger}$&
70.78&  74.5& 70.07& \textbf{110.75}& 109.77& \textbf{110.09}\\
\midrule
{MuJoCo Average}&   
 72.71& 73.28& 71.49& 69.09& \textbf{\color{blue}78.75}& \textbf{\color{blue}79.27}&
 45.87& 51.85& 51.77& 71.76& \textbf{\color{blue}79.53}& \textbf{\color{blue}81.12}$^{\dagger}$\\
\midrule
Pen-H&  
57.20 &   57.26&  62.33 & 66.07 & {\color{blue}72.75}&  \textbf{\color{blue}88.34}$^{\dagger}$ &
-3.15&  -3.71&  -0.99&  -2.81& -3.90& -3.70\\
Pen-C&   
57.71 &   62.94&  60.47 & 62.26& {\color{blue}69.05}&  \textbf{\color{blue}70.42}$^{\dagger}$ &
 5.30&  6.71&  3.06&  \textbf{19.13}& 15.03& 2.30\\
\midrule
Door-H&    
1.69  &   \textbf{5.05}$^{\dagger}$&  3.05  &  3.22 & \textbf{5.97}$^{\dagger}$& 3.67& 
-0.26&  -0.33&  -0.31&  -0.32& -0.33& -0.33\\
Door-C&   
-0.07 &   -0.10&  -0.09 &  -0.10& -0.07&  \textbf{{\color{blue}0.57}}$^{\dagger}$ &
-0.33&  -0.34&  -0.33&  -0.34& -0.33& -0.33\\
\midrule
Hammer-H&  
1.24  &   1.03&  1.50  & 0.54 & 1.26&  \textbf{\color{blue}6.63}$^{\dagger}$ &
0.86&  0.46&  0.75&  1.02&  0.89& 0.89\\
Hammer-C& 
\textbf{3.28}$^{\dagger}$ &   0.67 &  0.67  & 0.73   & 2.14&  \textbf{4.23}$^{\dagger}$  & 
0.34&  0.45&  0.25&  0.35&   0.24& 0.24\\
\midrule
{Adroit Average}& 
20.09 & 21.14 & 21.32 & 22.13 & {\color{blue}25.18}& \textbf{\color{blue}28.97}$^{\dagger}$&
0.46& 0.54& 0.40& \textbf{2.83}& 1.93& -0.15\\
\bottomrule 
\end{tabular}
}
\caption{The performance of DPR and C-DPR with CS label on Gym-MuJoCo and Adroit. For the MuJoCo task, we select three different datasets: medium (-M), medium-replay (-MR) and medium-expert (-ME). For Adroit tasks, we choose two different levels of datasets: human (-H) and cloned (-C). 
We use MLP* and TFM* to represent the recorded results of Uni-RLHF~\cite{yuanuni}. In the single algorithm, we use {\color{blue}blue} to indicate the improvement of the proposed method compared to previous methods and \textbf{bold} to indicate the best results. In all algorithms, we use $^{\dagger}$ to indicate the best results.}
\label{result:table_cs_all}
\end{table*}

\section{Related Work}
\label{related work}

\noindent \textbf{Offline Reinforcement Learning.} \
Offline reinforcement learning (RL) reduces extrapolation errors by limiting exploration without interacting with the environment~\cite{offline2020levine, 10078377}. 
Existing methods fall into two main categories: policy regularization and critic penalty. Policy regularization constrains the target policy to stay close to behavior policies or offline data~\cite{offpolicy2019fujimoto,a2021fujimoti}. 
Leveraging Transformers’ strength in sequence modeling, some approaches generate future actions from offline data using sequence information~\cite{NEURIPS2021_7f489f64, janner2021offline}. 
Critic penalty methods, on the other hand, reduce extrapolation errors by penalizing out-of-distribution data~\cite{conservative2020kumar, offline20221brandfonbrener,mao2024doubly}. 
Recent works further restrict value function learning to in-distribution data, limiting exclusively within the confines of the dataset~\cite{offline2022Ilya, garg2023extreme, xu2023offline, mao2024diffusion}.

\noindent \textbf{Offline Preference-based Reinforcement Learning.} \
Preference-based Reinforcement Learning 
(PbRL)~\cite{christiano2017deep, shin2021offline}, also referred to as RLHF, has gained wide attention~\cite{NEURIPS2022_b1efde53, dai2024safe, rafailov2024direct, azar2024general}. 
Offline PbRL emphasizes on incorporating preference information into offline reinforcement learning algorithms~\cite{kim2023preference, choi2024listwise, tu2024dataset}.
HPL~\cite{gao2025hindsight} models human preferences using the future outcomes of trajectory segments, i.e., hindsight information.
FTB~\cite{zhang2023flow} utilizes diffusion models to continuously generate expert-level data based on preference trajectories and performs behavior cloning using this expert information. 
In addition, some work~\cite{knox2024models, hejna2024contrastive} uses segment regret to model human preferences and guide reward learning.
OPPO~\cite{kang2023beyond} and DPPO~\cite{NEURIPS2023_de8bd6b2} eliminate direct reward function modeling, instead aligning policy with preference information through reinforcement learning. 
Meanwhile, IPL~\cite{hejna2024inverse} extends the DPO~\cite{rafailov2024direct} algorithm to optimize the value function by incorporating preference information.

\noindent \textbf{Diffusion-based Reinforcement Learning.} \
The diffusion model is a powerful generative tool widely used in reinforcement learning. Diffuser~\cite{pmlr-v162-janner22a} uses it to generate optimal trajectories in offline settings.
Building on this, Decision Diffusion~\cite{ajay2023is} 
adds an inverse dynamics model to output optimal actions. Compared to trajectory generation, policy modeling is more efficient and improves policies more directly~\cite{wang2023diffusion, NEURIPS2023_d45e0bfb, pmlr-v202-lu23d, chi2023diffusion}.
Recent studies~\cite{wang2024diffail, lai2024diffusion, chen2024diffusion} show that diffusion models can not only generate trajectories and model policies, but also act as discriminators to help learn target policies.

\section{Experiments}
\label{experiments}

In this section, we evaluate the performance of DPR and C-DPR by answering the following questions:
\begin{itemize}
    \item Do diffusion model-derived rewards enhance offline RL with human preference over other methods?
    \item How do DPR and C-DPR compare to other offline PbRL approaches?
    \item How does the acquisition of diffusion reward relate to the diffusion step size and the size of the preference dataset?
\end{itemize}

\noindent\textbf{Setup.} 
We mainly choose the Uni-RLHF~\cite{yuanuni} dataset as the evaluation dataset, which extends D4RL~\cite{fu2020d4rl} with preference information comprising both crowd-sourced~(\textbf{CS}) human labels and scripted teacher~(\textbf{ST}) labels based on ground truth rewards. 
Here we mainly consider two environments: Gym-MuJoCo and Adroit.
Following Uni-RLHF, we evaluate the performance of the reward functions applied to offline reinforcement learning algorithms through $\mathrm{ normalized\ score}=\frac{\mathrm{score-score(random)}}{\mathrm{score(expert)-score(random)}} * 100$, where $\mathrm{score(random)}$ and $\mathrm{score(expert)}$ are preset minimum and maximum values.
We provide a comprehensive introduction to our experiment setup in Appendix A.

For the baseline of reward acquisition methods, we choose two common reward model modeling methods, \textbf{MLP}-based Markovian reward and Transformer-based(\textbf{TFM}) non-Markovian reward, and evaluate them on existing offline RL algorithms CQL~\cite{conservative2020kumar}, TD3BC~\cite{a2021fujimoti}, IQL~\cite{offline2022Ilya}. 
In addition, we choose DTR~\cite{tu2024dataset}, which performs offline PbRL using the MLP-based Markovian reward model, as our baseline for comparison of reward acquisition effects.
For the baseline of offline PbRL, we consider the following:
\begin{itemize}
    \item DPPO~\cite{NEURIPS2023_de8bd6b2}: direct optimization policy through preference information;
    \item Preference Transformer (PT)~\cite{kim2023preference}: building a weighted btmodel through transformer;
    \item IPL~\cite{hejna2024inverse}: ignoring the learning of the reward function by adding preference information to optimize the value function;
    \item FTB~\cite{zhang2023flow}: generating high-quality data by a conditional diffusion model and learning policy through behavior cloning.
\end{itemize}
Consistent with previous work, we randomly select 5 seeds and evaluate 10 trajectories for each task, using the mean and standard deviation as the reported performance. 


\begin{figure}[ht]
    \centering
    \includegraphics[width=1.0\linewidth]{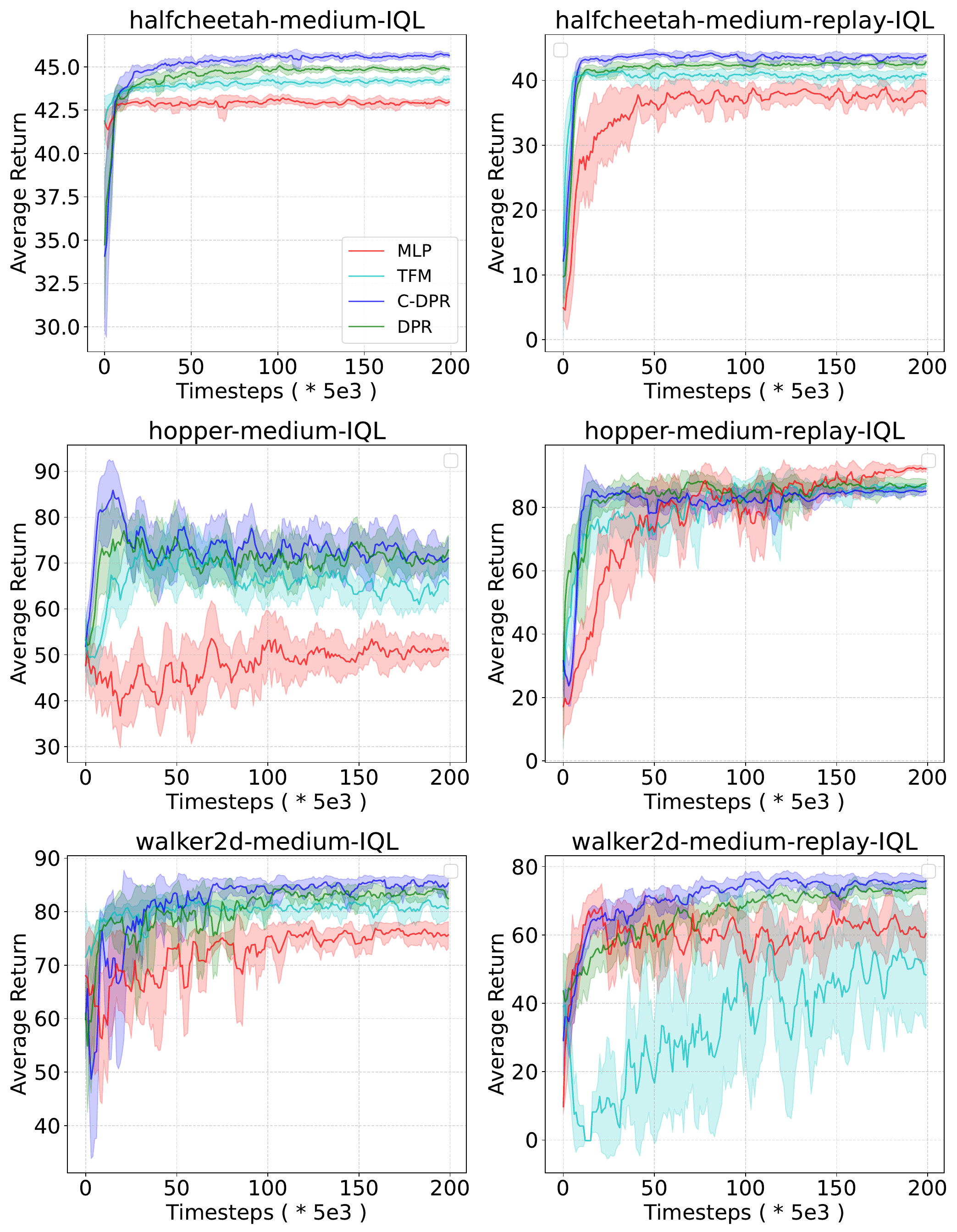}
    \caption{Training curve of IQL algorithm based on CS-label.}
    \label{fig:result_cs_iql}
\end{figure}

\begin{figure}[ht]
    \centering
    \subfigure[DPR]{\includegraphics[width=0.9\linewidth]{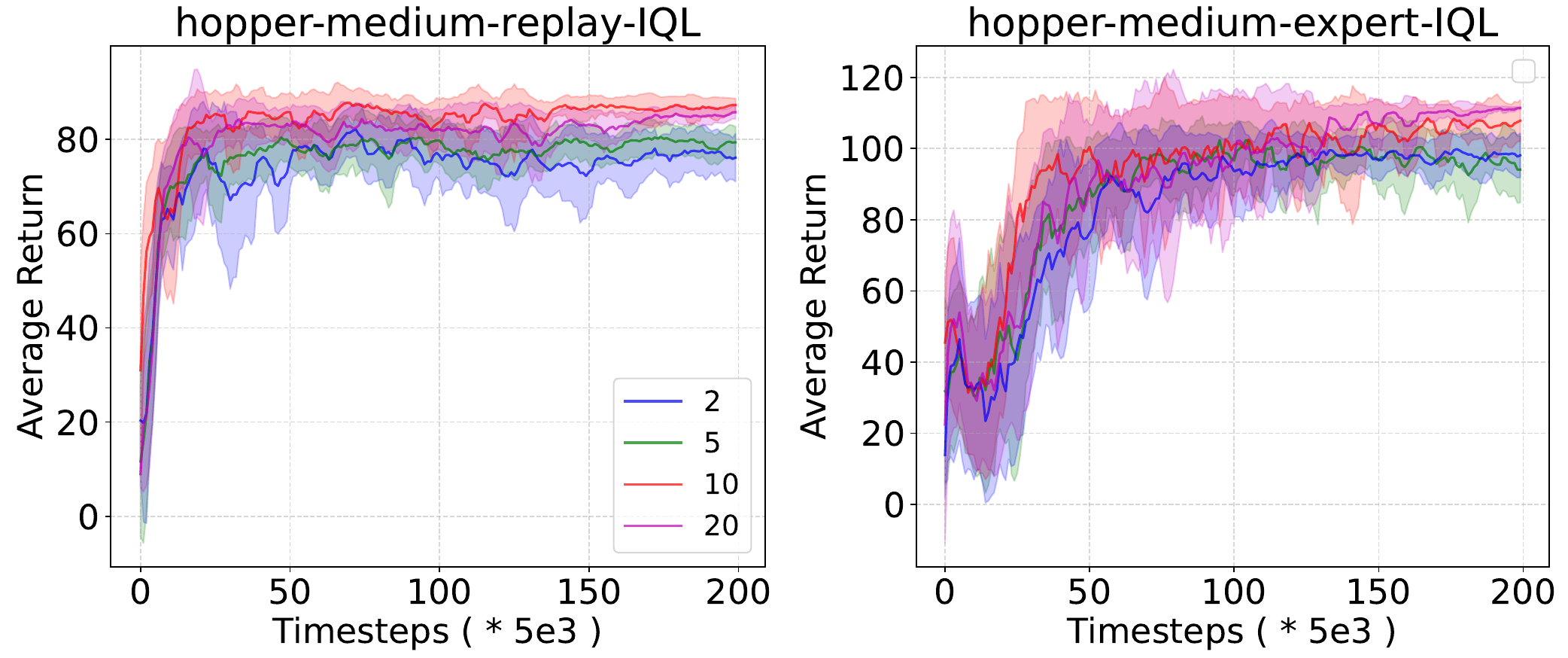}}
    
    \subfigure[C-DPR]{\includegraphics[width=0.9\linewidth]{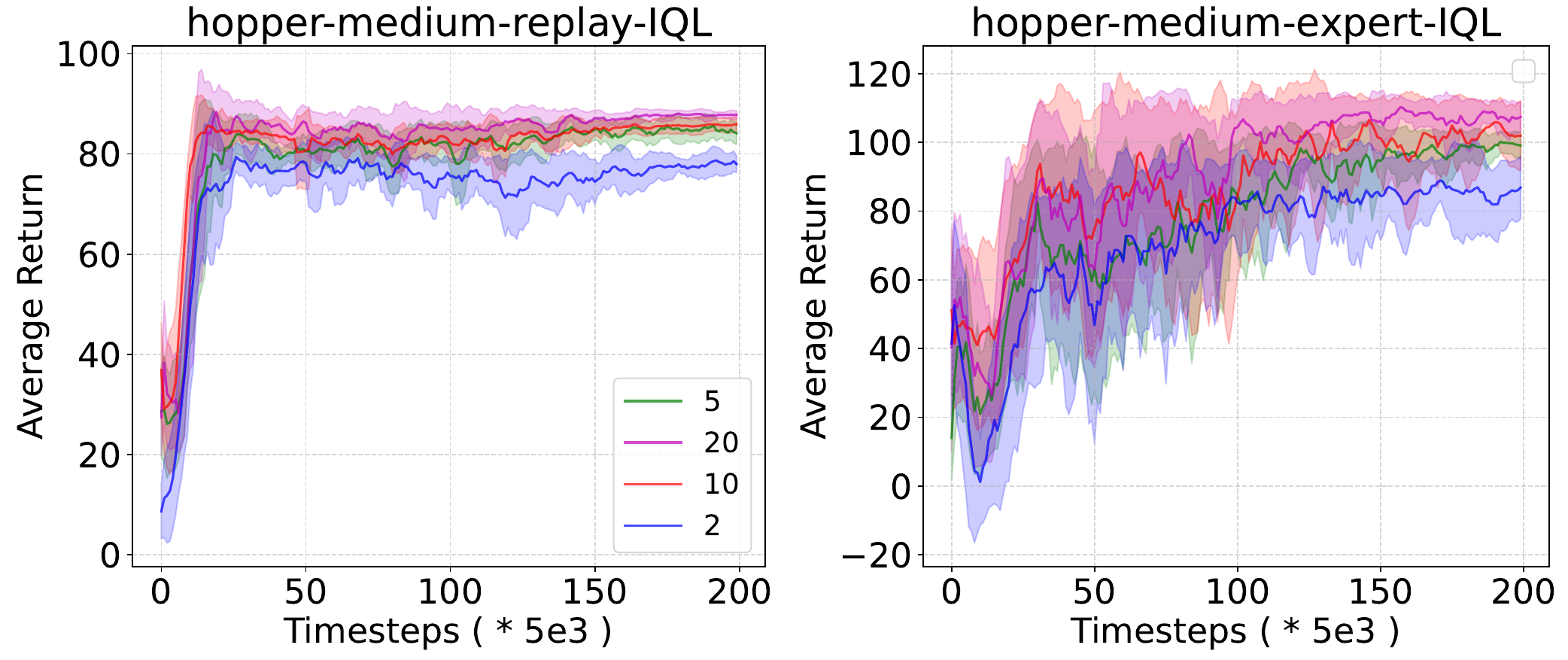}}
    \caption{Training curve of diffusion timestep.}
    \label{fig:result_diffusion_timestep}
\end{figure}


\begin{table*}[t]
\begin{center}
\small
\setlength{\tabcolsep}{7pt}{
\begin{tabular}{c|cc|cc|cc|cccc}
\toprule  
\multirow{2}*{Dataset}&\multicolumn{2}{|c}{CQL}&  \multicolumn{2}{|c}{IQL}& \multicolumn{2}{|c|}{TD3BC}& \multirow{2}*{FTB}& \multirow{2}*{DPPO}& \multirow{2}*{PT}& \multirow{2}*{IPL}\\
\cmidrule{2-7} 
~ & DPR& C-DPR& DPR& C-DPR& DPR& C-DPR& ~& ~& ~& ~\\
\midrule  
HalfCheetah-MR&   42.67&  43.62&  42.29&	43.88&	40.35&	42.87&	39.0& 40.8& \textbf{44.4}& 42.5\\
HalfCheetah-ME&   94.3&	  \textbf{95.49}&  88.06&	87.44&	93.63&  72.21&	91.3& 92.6& 87.5& 87.0\\
\midrule
Hopper-MR&    75.98&	87.88&	84.91&	86.48&	75.49&	88.84&	\textbf{90.8}&  73.2&  84.5&  73.6\\
Hopper-ME&    73.98&	74.24&	104.36&	102.64&	104.04&	102.84&	\textbf{110.0}& 107.2& 69.0&  74.5\\
\midrule
Walker2D-MR&   62.97&	79.63&	71.37&	86.01&	74.13&	\textbf{89.45}&	79.9& 50.9& 71.2& 60.0\\
Walker2D-ME&   109.33&	109.55&	110.87&	\textbf{112.37}&	109.98& 110.57&	109.1& 108.6& 110.1& 108.5\\
\midrule
{MuJoCo Average}&   76.55& 81.73& 83.64& \textbf{86.47}& 82.93& 84.46& \textbf{86.68}& 78.88& 77.78& 74.35\\
\bottomrule 
\end{tabular}
}
\caption{Comparison results with offline PbRL.}
\label{result:table_PbRL}
\end{center}
\end{table*}


\begin{table}[t]
\begin{center}
\small
\setlength{\tabcolsep}{1mm}{
\begin{tabular}{c|ccc|ccc}
\toprule  
\multirow{2}*{Dataset}& \multicolumn{3}{|c}{DPR+IQL}& \multicolumn{3}{|c}{C-DPR+IQL}\\
\cmidrule{2-7} 
~ & (2)& (50)& (2000)& (2)& (50)& (2000)\\
\midrule  
HalfCheetah-M&    42.7&	\textbf{44.3}&	\textbf{44.9}&	42.6& \textbf{47.3}&	45.5\\
HalfCheetah-MR&   36.6&	41.0&	\textbf{42.7}&	37.2& 42.9&	\textbf{44.0}\\
HalfCheetah-ME&   60.5&	\textbf{91.1}&  87.9&	58.8& 91.0&   \textbf{93.6}\\
\midrule
Hopper-M&        72.5&	70.1&	\textbf{74.4}&	 74.7&	\textbf{76.9}&	70.9\\
Hopper-MR&       33.8&	78.2&	\textbf{87.5}&	 32.8&	\textbf{89.5}&	86.1\\
Hopper-ME&       96.4&	104.0&	\textbf{108.4}& 88.0&	\textbf{107.9}&	99.3\\
\midrule
Walker2D-M&      75.2&	78.7&	\textbf{83.4}&	76.6&	78.15&	\textbf{85.7}\\
Walker2D-MR&     56.1&	\textbf{72.3}&	67.2&	64.6&	\textbf{85.2}&	75.9\\
Walker2D-ME&     109.1&\textbf{111.8}&	\textbf{111.8}&	109.7&	\textbf{112.3}& \textbf{111.9}\\
\midrule 
MuJoCo Average&  64.8& 76.8&   \textbf{78.7}&  65.0&  \textbf{81.2}& 79.2\\
\bottomrule 
\end{tabular}
}
\caption{Ablation experiments with different batch sizes of preference data. We use \textbf{bold} to indicate the best results of DPR and C-DPR, respectively.}
\label{result:table_batchsize}
\end{center}
\end{table}

\subsection{Evaluation Results with Human Labels}
\label{experiment:Performance_CS}

We assess the capability of rewards derived from diffusion models to facilitate downstream offline RL algorithms, in comparison to MLP-based Markovian reward and Transformer-based non-Markovian reward.
We list some results with DPR and C-DPR under CS-labels in Table~\ref{result:table_cs_all} and Figure~\ref{fig:result_cs_iql}.
Appendix B provides the complete experimental results, including comparisons with CQL algorithm and \textit{oracle} rewards, all reported with means and standard deviations.

Based on the above results, it can be seen that for Gym-MuJoCo tasks, the rewards obtained through diffusion models demonstrate remarkable advantages, regardless of the downstream algorithm. For Adroit tasks, despite some poor performances due to improper downstream algorithm choices, we remain confident in using DPR or C-DPR for offline RL algorithms like IQL, which perform well. In addition, the performance of C-DPR is better than that of DPR, which indicates the significance of trajectory-wise preference labels.
Furthermore, surprisingly, DPR and C-DPR outperform \textit{oracle} reward on some datasets, suggesting that manually designed reward functions may not be optimal for offline policy learning.
We also compared DTR with and without C-DPR (Table~\ref{appendix:table_DTR_all} in Appendix B). Even for offline PbRL methods, using a conditional diffusion model for reward inference can improve performance.

Following some offline PbRL algorithms, we also consider the preference labels from Preference Transformer~\cite{kim2023preference} and evaluate our method on them, as shown in Table~\ref{result:table_PbRL}.
The results demonstrate that our method performs well and is comparable to FTB. 
This is attributed to our method to obtain suitable rewards using the diffusion model and directly apply them to downstream offline RL algorithms.

To evaluate generalization, we apply diffusion rewards from offline datasets to the online RL algorithm PPO and compare them with MLP-based Markovian rewards. As shown in Table~\ref{result:table_ppo_all} in Appendix B, our method achieves better performance, even in the online scenario.

\subsection{Ablation Study of Diffusion Reward Model} \label{experiment:Performance_timestep}


We first conduct ablation studies to explore how diffusion step size affects model performance, with results shown in Figure~\ref{fig:result_diffusion_timestep} and more details in Appendix B.3. 
Upon examining these results, it can be seen that as the diffusion step size increases, the performance of downstream algorithms gradually improves, aligning with our comprehension of diffusion models. 
Despite achieving good and comparable performance with diffusion step sizes of both 10 and 20, we choose 10 steps due to the higher cost associated with multi-step sampling.
Additionally, Appendix.Table~\ref{result:table_noiseless_all} compares performance before and after adding noise, showing that noise-perturbed data improve downstream algorithms—likely because diffusion models provide more robust reward signals.

To study how the size of the preference dataset affects the diffusion reward model, we conduct an ablation experiment using different numbers of preference trajectory pairs. We randomly selected 
$K$ pairs of preference trajectories (each trajectory considering $200$ pairs) to train the diffusion model and evaluate it with the downstream IQL algorithm. 
As demonstrated in Table~\ref{result:table_batchsize}, competitive performance can be achieved with just $2$ preference pairs, and using $50$ pairs achieves results on par with utilizing the full set of $2000$. Notably, increasing more preference data does not always lead to further improvements, as fuzzy preference annotations may hinder the effectiveness of the diffusion model.



\subsection{Diffusion Preference-based Reward with ST Labels}

To evaluate diffusion-based preference rewards against oracle rewards, we extend experiments with DPR and C-DPR on ST labels. Appendix B shows that DPR and C-DPR significantly improve performance, even with oracle preferences.


\section{Conclusion and Discussion}
\label{conclusion and discussion}

In this paper, we propose two diffusion-based reward mechanisms: DPR and C-DPR, which discriminatively derive step-wise rewards through the robust classification capabilities of diffusion models. 
Unlike the Bradley-Terry model, DPR and C-DPR extract step-wise rewards from trajectory-wise preferences, alleviating the effect of the preference trajectory on step-wise reward.
In addition, they can be seamlessly integrated into standard offline RL algorithms, providing suitable reward support. 
The results of our experiment validate the effectiveness of the diffusion-based rewards.
Although the rewards obtained from diffusion models yield promising results, naively dividing data by quality can easily cause bias. Moreover, the underlying reasons why diffusion-based rewards outperform those from the traditional Bradley-Terry model remain unclear. Exploring a potential combination of the two approaches also presents a promising direction.

\bibliography{reference}


\appendix
\onecolumn

\section{Appendix}

\section{A. Environment and Experiment Details}

\label{appendix:detail}

Our main experiments are based on the open-source code Uni-RLHF~\cite{yuanuni}\footnote{https://github.com/pickxiguapi/Uni-RLHF-Platform}, and the data involved in the experiments is sourced from the D4RL~\cite{fu2020d4rl} \footnote{https://github.com/Farama-Foundation/D4RL} dataset. Following some offline PbRL algorithms, we also consider the preference labels from Preference Transformer~\cite{kim2023preference}. For PPO~\cite{schulman2017proximal} algorithm, we choose open-source code base Stable Baselines3\footnote{https://github.com/DLR-RM/stable-baselines3}. 
Here we mainly describe two environments in D4RL: Gym-MuJoCo and Adroit.

\subsection{A.1 Environment}
\label{appendix:environment}

\noindent \textbf{Gym-MuJoCo} \
In MuJoCo-based localization tasks, the core focus is on precise control of the robot to achieve smooth and reasonable motion. This covers three different task scenarios: HalfCheetah, Hopper, and Walker2D. 
For each task, we regard the data gathered by the policy learned through the online RL algorithm SAC~\cite{pmlr-v80-haarnoja18b} as expert data. Depending on various data collection approaches, we consider three types of datasets: using the expert policy trained to 1/3 performance (medium), using data before training to medium performance (medium-replay), and a mixture of medium and expert data (medium-expert). In the experiment, we use the `-v2' version.

\noindent \textbf{Adroit} \
Adroit task controls a 24-DoF robotic arm to hammer a nail (Hammer), open doors (Door), or turn pen(Pen).
Based on the above tasks, we consider two types of datasets: small-scale datasets obtained through human demonstrations (Human) and cloned datasets that mimic and mix human demonstrations (Cloned). In the experiment, we use the `-v1' version.

\subsection{A.2 Preference Label}
\label{appendix:preference_label}
For the preference label, we primarily rely on crowdsourcing data and scripted teacher data provided by Uni-RLHF. For a complete understanding of the data content, please consult Uni-RLHF. 
Meanwhile, we also consider partial preference labels from the Preference Transformer~\cite{kim2023preference} in order to enable comparison with other baselines.

\subsection{A.3 Experiment Detail}
\label{appendix:experiment_detail}

For the diffusion model, we present all hyperparameter settings in Table~\ref{appendix:table_hyperparmeter}. 
Note that C-DPR introduces an additional hyperparameter, Condition Dimension, on top of the hyperparameters used in DPR.

\begin{table*}[h]
\begin{center}
\setlength{\tabcolsep}{7pt}{
\begin{tabular}{ccc}
\toprule  
Algorithm& Hyperparameter& Value\\
\midrule  
\multirow{9}{*}{DPR}&
  Model Architecture& MLP\\
& Hidden Layer& 4\\
& Hidden Dimension& 128\\
& Timestep&    10 \\
& Noise Schedule&         linear\\
& Batch Size&  256\\
& Optimizer&   Adam\\
& Learning Rate& $3\times10^{-4}$\\
& Training Epoch& 500\\
\midrule
C-DPR& Condition Dimension& 10\\
\bottomrule 
\end{tabular}}
\caption{Hyperparameter for diffusion model}
\label{appendix:table_hyperparmeter}
\end{center}
\end{table*}

For downstream offline RL algorithms, a specific implementation can refer to Uni-RLHF. For the implementation of the DTR~\cite{tu2024dataset}, please refer to https://github.com/TU2021/DTR.

For each experiment, we perform 1,000,000 gradient update steps, running for 8 hours on a single RTX 4090 24GB GPU.

\section{B. Full Experiment Results}
\label{appendix:result_all}

\begin{figure*}[ht]
    \centering
    \includegraphics[width=0.5\linewidth]{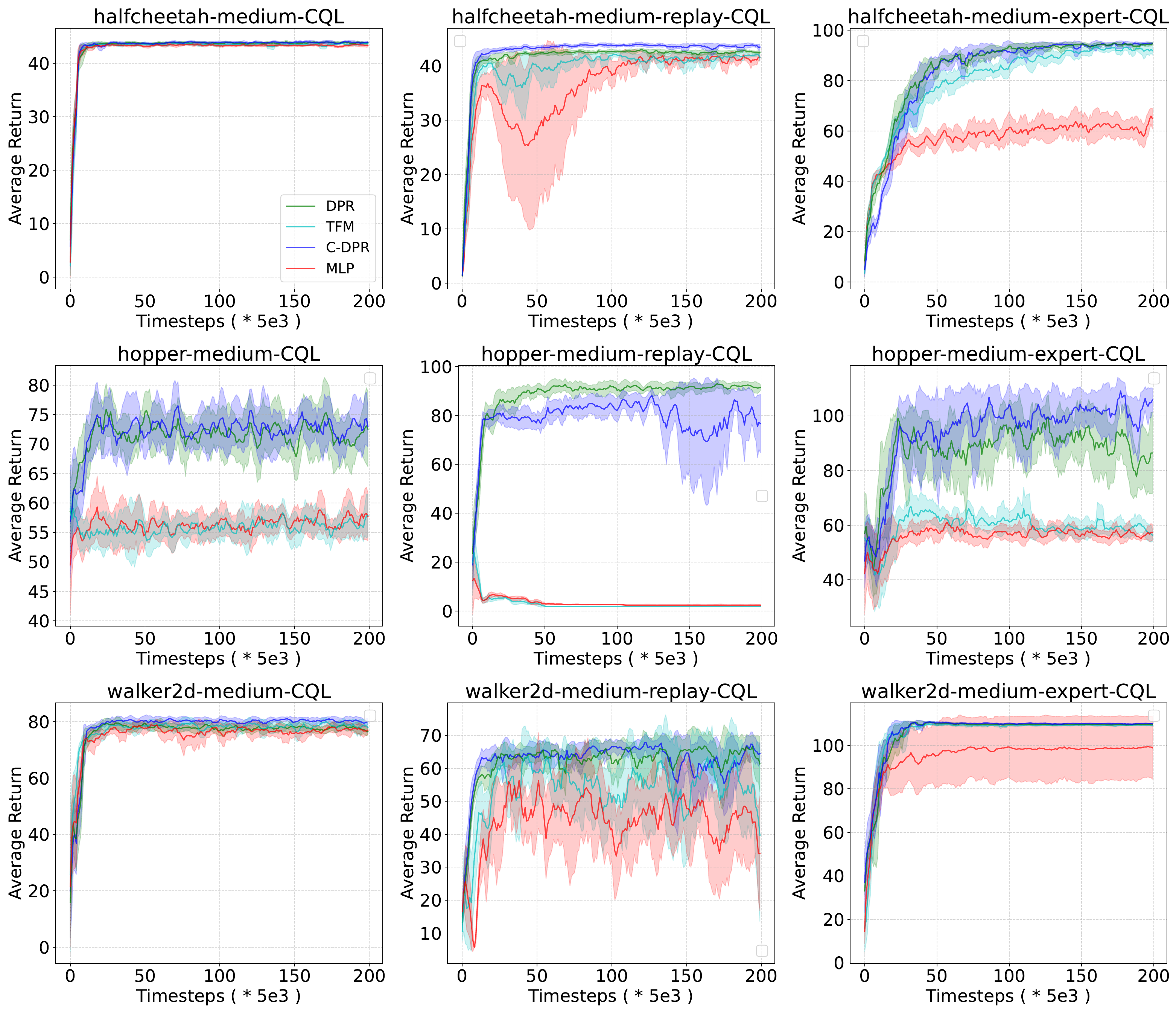}
    \caption{Training curve of CQL algorithm based on CS-label}
    \label{fig:result_cql_cs}
\end{figure*}

\begin{figure*}[ht]
    \centering
    \includegraphics[width=0.5\linewidth]{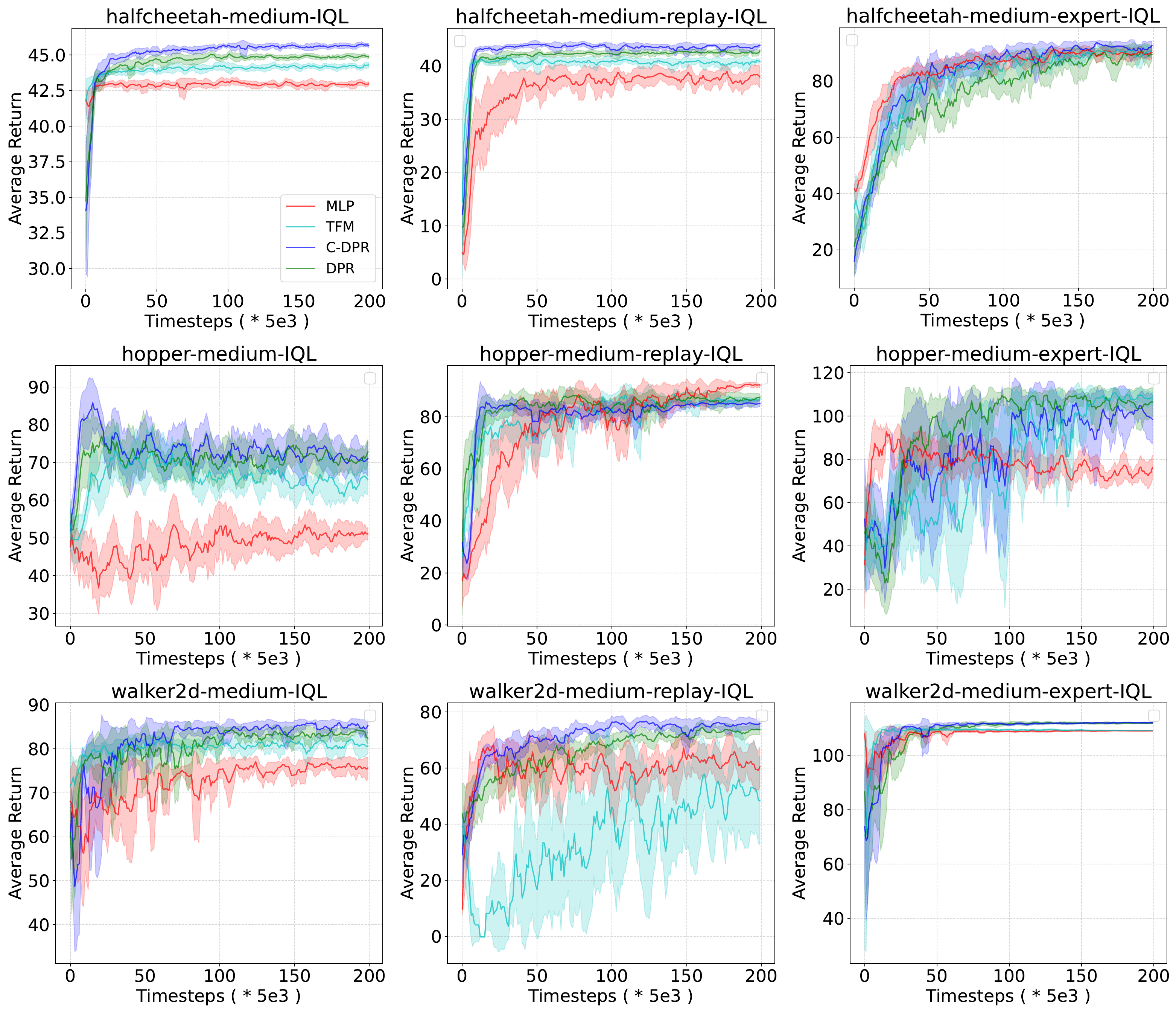}
    \caption{Training curve of IQL algorithm based on CS-label}
    \label{fig:result_iql_cs}
\end{figure*}

\begin{figure*}[ht]
    \centering
    \includegraphics[width=0.65\linewidth]{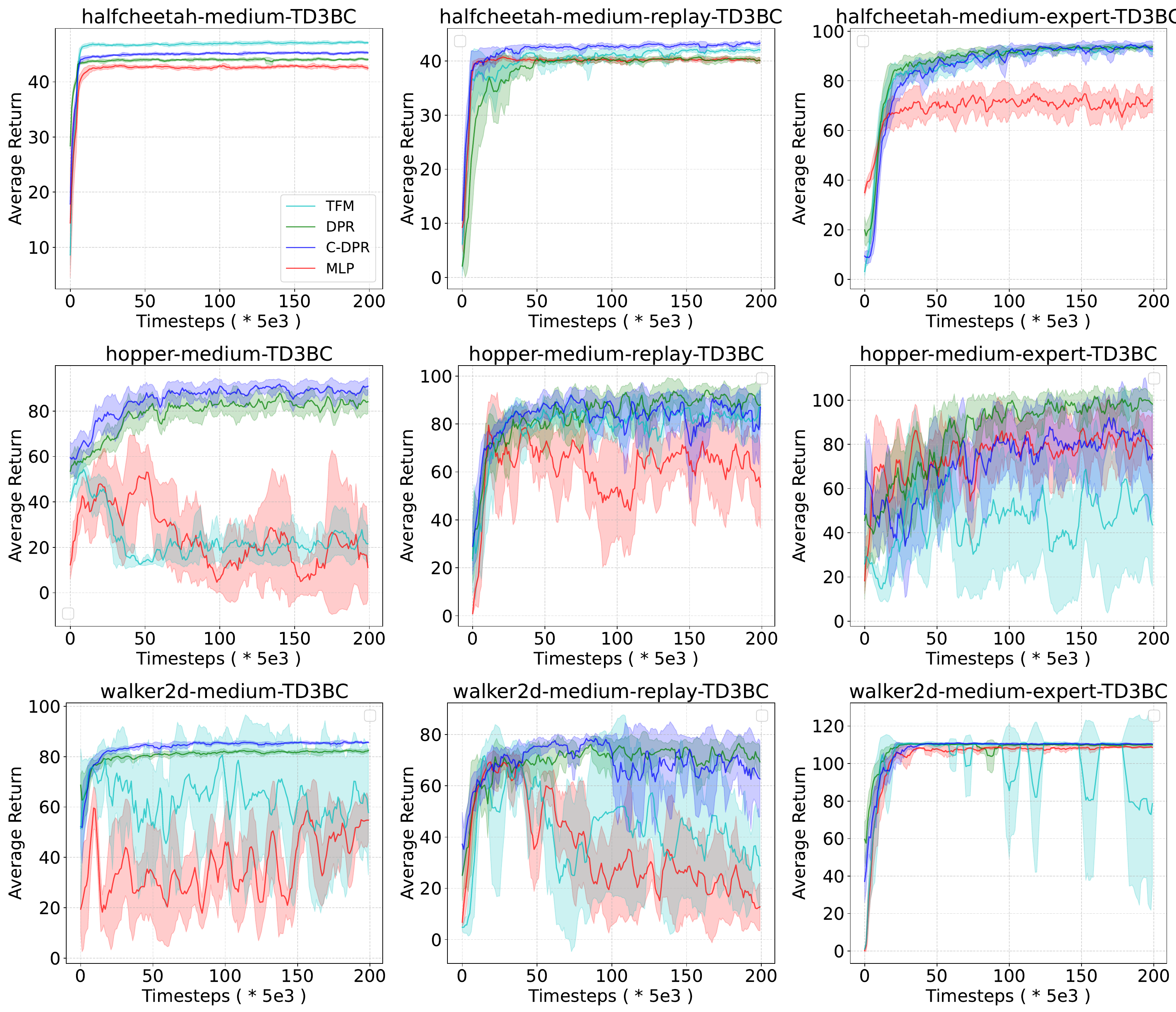}
    \caption{Training curve of TD3BC algorithm based on CS-label}
    \label{fig:result_td3bc_cs}
\end{figure*}

\begin{figure*}[ht]
    \centering
    \includegraphics[width=0.65\linewidth]{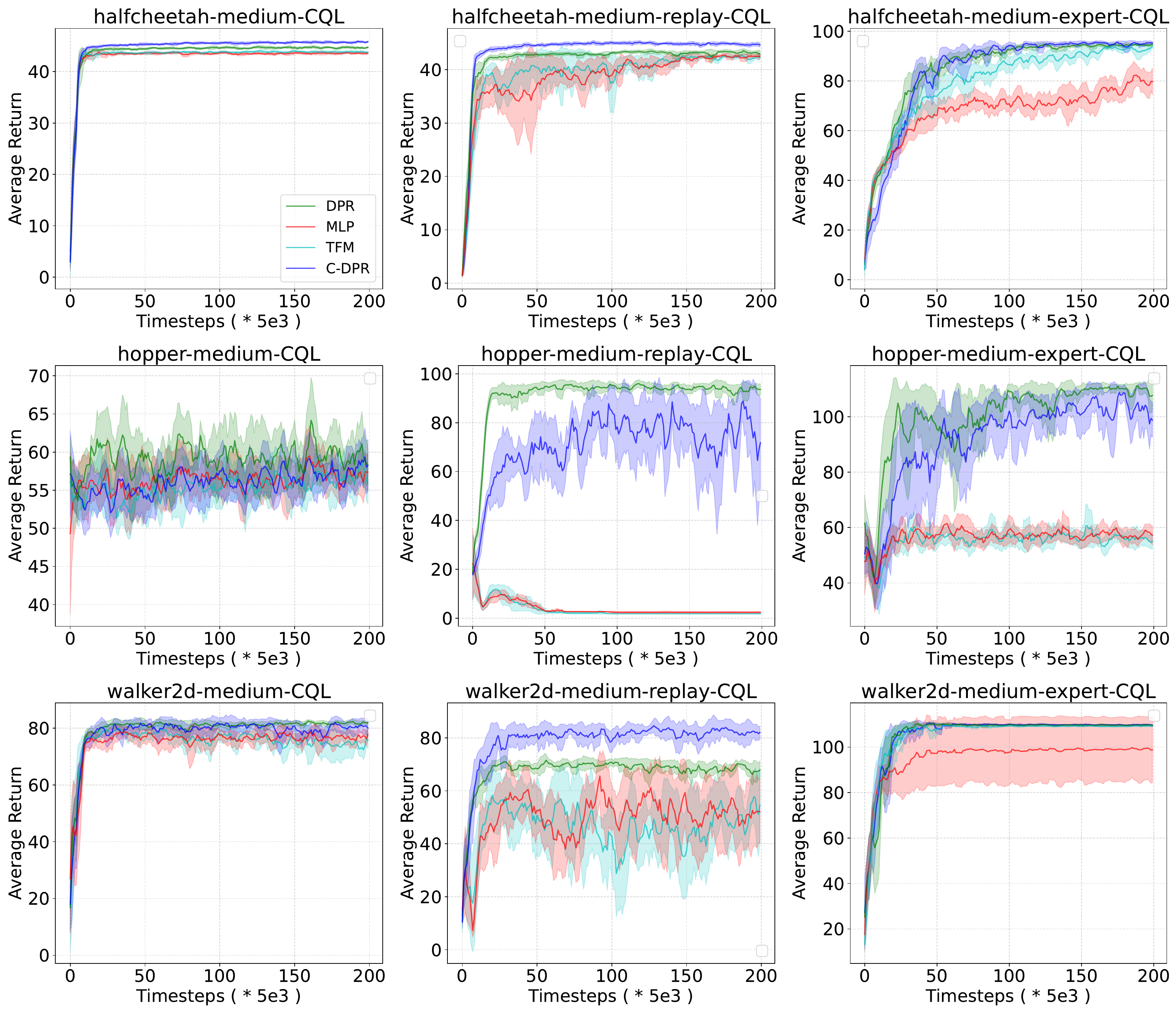}
    \caption{Training curve of CQL algorithm based on ST-label}
    \label{fig:result_cql_st2}
\end{figure*}

\begin{figure*}[ht]
    \centering
    \includegraphics[width=0.65\linewidth]{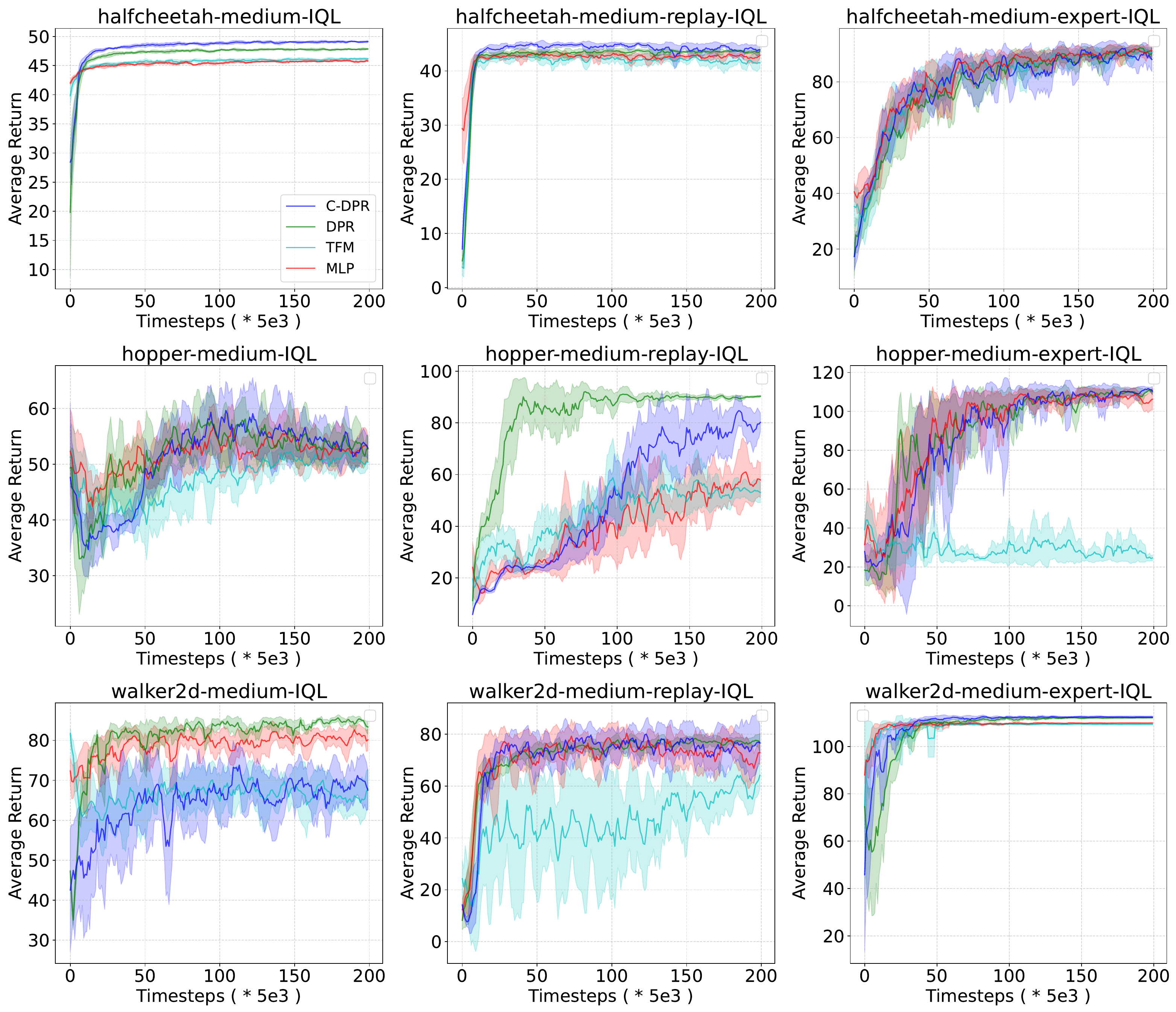}
    \caption{Training curve of IQL algorithm based on ST-label}
    \label{fig:result_iql_st}
\end{figure*}

\begin{figure*}[ht]
    \centering
    \includegraphics[width=0.65\linewidth]{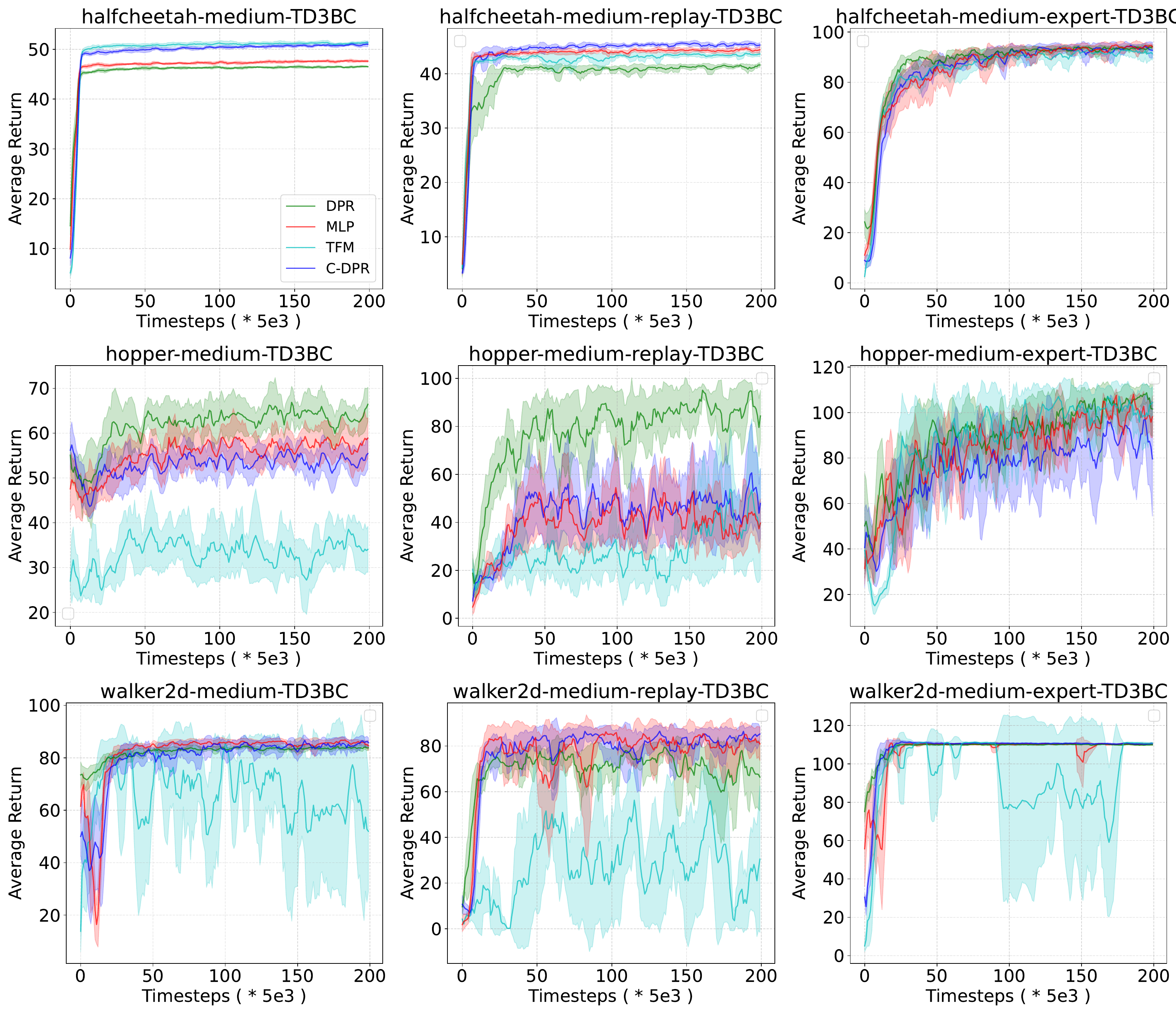}
    \caption{Training curve of TD3BC algorithm based on ST-label}
    \label{fig:result_td3bc_st}
\end{figure*}

We record the experiment results of all reward methods, including CS-label and ST-label, on Gym-MuJoCo and Adroit. Specifically, Table~\ref{result:table_cql_cs},~\ref{result:table_iql_cs} and ~\ref{result:table_td3bc_cs} detail the CS-label results of CQL, IQL, and TD3BC algorithms, while Table~\ref{result:table_cql_ST},~\ref{result:table_iql_ST} and ~\ref{result:table_td3bc_ST} record the ST-label results of the aforementioned algorithms, respectively. Furthermore, we depict the training curves of the three algorithms in CS-label in Figure~\ref{fig:result_cql_cs}, ~\ref{fig:result_iql_cs} and ~\ref{fig:result_td3bc_cs}, while the ST-label curves for are shown in Figure~\ref{fig:result_cql_st2},~\ref{fig:result_iql_st} and~\ref{fig:result_td3bc_st}. We also present the complete CS-label comparison results of DTR and offline PbRL in Table~\ref{appendix:table_DTR_all} and Table~\ref{result:table_PbRL_all}, respectively. 
In addition, Table~\ref{result:table_ppo_all} presents the performance of the PPO algorithm in online settings, while Tables~\ref{result:table_batchsize_all} and~\ref{result:table_noiseless_all} report the ablation results with respect to varying numbers of preferences and the inclusion or exclusion of the noise disturbance, respectively.

\subsection{B.1 Performance of Offline PbRL}
\label{appendix:result_table_all}

\begin{table*}[h]
\begin{center}
\small
\setlength{\tabcolsep}{4pt}{
\begin{tabular}{c|c|cccc|cc}
\toprule  
Dataset& Oracle& MLP& MLP(paper)& TFM& TFM(paper)& DPR& C-DPR\\
\midrule  
HalfCheetah-M&   47.04 ± 0.22&        43.25±0.21 &  43.4±0.1&  43.54±0.11 & 43.5±0.24&  \textbf{44.15±0.22}&  43.95±0.16 \\
HalfCheetah-MR&  45.04 ± 0.27& 42.06±0.14 &  41.9±0.1&  37.94±4.83 & 40.97±1.48&  42.60±0.24&  \textbf{{\color{blue}43.80±0.58}} \\
HalfCheetah-ME&  95.63 ± 0.42& 59.53±7.88 &  62.7±7.1&  91.47±2.68 & 64.86±8.6&  \textbf{{\color{blue}94.94±0.51}}& \textbf{{\color{blue}94.89±0.25}} \\
\midrule
Hopper-M&     59.08 ± 3.77&        58.07±2.24 &  54.7±3.4&  56.58±2.23 & 63.47±5.82&  \textbf{{\color{blue}73.05±4.18}}&  {\color{blue}70.35±3.25} \\
Hopper-MR&    95.11 ± 5.27& 2.30±0.40  &  1.8±0.0&   1.81±0.006  & 52.97±13.04&  \textbf{{\color{blue}90.16±1.04}}&  {\color{blue}82.55±9.59} \\
Hopper-ME&    99.26 ± 10.91& 57.79±3.27 &  57.4±4.9&  57.19±1.78 & 57.05±4.83&  {\color{blue}87.64±19.00}& \textbf{{\color{blue}106.24±9.08}}\\
\midrule 
Walker2D-M&   80.75 ± 3.28&        77.97±1.588 &  76.0±0.9&  77.15±2.72 & 77.22±2.44&  77.00±1.82&  \textbf{{\color{blue}80.30±1.17}} \\
Walker2D-MR&  73.09 ± 13.22& 28.31±27.56 &  20.6±36.3&  38.18±31.17 & 1.82±0.00 &  {\color{blue}60.67±8.88}&  \textbf{{\color{blue}66.96±2.42}} \\
Walker2D-ME&  109.56 ± 0.39& 102.30±13.36&  92.8±22.4&  108.90±0.43& 98.96±17.1&  \textbf{109.31±0.37}& \textbf{109.73±0.47}\\
\midrule
{MuJoCo Average}&  78.28& 52.40& 50.44&  57.01& 55.64& {\color{blue}75.50}& 
 \textbf{{\color{blue}77.64}}\\
\midrule
Pen-H&      13.71 ± 16.98&  15.40±20.49&  6.53±11.79&  17.45±16.87&  \textbf{23.77±14.23}& 15.69±11.69& 17.64±13.69\\
Pen-C&      1.04 ± 6.62&    \textbf{6.05±10.89}&  2.88±8.42 &  2.98±7.24&  3.18±7.55& 3.92±10.37& \textbf{5.96±5.25}\\
\midrule
Door-H&     5.53 ± 1.31&    6.12±1.88&  \textbf{10.31±6.46}&  3.62±2.76&  8.81±3.14  & 3.75±1.53& 3.07±2.75\\
Door-C&     -0.33 ± 0.01&   -0.33±0.00&  -0.34±0.00&  -0.33±0.00&  -0.34±0.00 & -0.33±0.00& -0.33±0.00\\
\midrule
Hammer-H&   0.14 ± 0.11&   1.02±0.47&  0.70±0.18 &  1.26±0.61&  1.13±0.34  & 1.33±1.52& 0.89±0.49\\
Hammer-C&   0.30 ± 0.01&   0.28±0.01&  0.28±0.01 &  0.27±0.01&  0.29±0.01  & 0.28±0.01& 0.29±0.00\\
\midrule
{Adroit Average}& 3.40& 4.75& 3.39&  4.20& \textbf{6.14}& 4.10& 4.58\\
\bottomrule 
\end{tabular}}
\caption{Offline Preference-based RL in CS-label with CQL algorithm}
\label{result:table_cql_cs}
\end{center}
\end{table*}

\begin{table*}[h]
\begin{center}
\small
\setlength{\tabcolsep}{4pt}{
\begin{tabular}{c|c|cccc|cc}
\toprule  
Dataset& Oracle & MLP& MLP(paper)& TFM& TFM(paper)& DPR& C-DPR\\
\midrule  
HalfCheetah-M&   48.31 ± 0.22&          43.01±0.24 & 43.3±0.2 &  44.24±0.2 & 43.24±0.34 & {\color{blue}44.90±0.12} &  \textbf{\color{blue}45.57±0.13} \\
HalfCheetah-MR&  44.46 ± 0.22&   36.90±2.08 & 38.0±2.3 &  40.83±0.79 & 39.49±0.41 & {\color{blue}42.79±0.32} & \textbf{\color{blue}44.09±0.29} \\
HalfCheetah-ME&  94.74 ± 0.52&   91.48±1.08 & 91.0±2.3 &  91.76±0.41 & 92.20±0.91 & 87.99±4.66 & \textbf{\color{blue}93.62±0.95} \\
\midrule
Hopper-M&      67.53 ± 3.78&          53.82±4.05 & 50.8±4.2 &  62.78±2.28 & 67.81±4.62 & \textbf{\color{blue}74.49±2.1} & {\color{blue}70.98±3.93} \\
Hopper-MR&     97.43 ± 6.39&   91.74±1.58 & 87.1±6.1 &  \textbf{88.77±2.25} & 22.65±0.8 & 87.53±1.06 & 86.10±1.47 \\
Hopper-ME&     107.42 ± 7.80&   81.73±4.34 & 94.3±7.4 &  105.10±3.87& \textbf{111.43±0.41}& 108.44±6.08& 99.34±10.07 \\
\midrule
Walker2D-M&   80.91 ± 3.17&          74.12±1.12 & 78.4±0.5 &  81.57±1.79 & 79.36±2.31 & {\color{blue}83.47±1.59} & \textbf{\color{blue}85.77±0.4} \\
Walker2D-MR&  82.15 ± 3.03&   69.58±7.92 & 67.3±4.8 &  56.92±17.06 & 56.52±4.62 & {\color{blue}67.23±8.51} & \textbf{\color{blue}75.95±1.72} \\
Walker2D-ME&  111.72 ± 0.86&   109.02±0.12& 109.4±0.1&  109.07±0.07& 109.12±0.14& \textbf{\color{blue}111.89±0.18}& \textbf{\color{blue}111.98±0.26}\\
\midrule
{MuJoCo Average}& 81.63& 72.71& 73.28& 71.49& 69.09& \textbf{\color{blue}78.75}& \textbf{\color{blue}79.27}\\
\midrule
Pen-H&      78.49 ± 8.21&  57.20±14.4 &   57.26±28.86 &  62.33±8.59 & 66.07±10.45  & {\color{blue}72.75±7.55}&  \textbf{\color{blue}88.34±10.38} \\
Pen-C&      83.42 ± 8.19&  57.71±17.44 &   62.94±23.44 &  60.47±13.52 & 62.26±20.86  & {\color{blue}69.05±13.88}&  \textbf{\color{blue}70.42±10.44} \\
\midrule
Door-H&     3.26 ± 1.83&   1.69±0.59  &   \textbf{5.05±4.34}  &  3.05±0.26  &  3.22±2.45   & \textbf{5.97±3.74}&  3.67±3.36  \\
Door-C&     3.07 ± 1.75&  -0.07±0.01 &   -0.10±0.02  &  -0.09±0.00 &  -0.01±0.08 & -0.07±0.01&  \textbf{\color{blue}0.57±0.94} \\
\midrule
Hammer-H&   1.79 ± 0.80&   1.24±0.41  &   1.03±0.13  &  1.50±1.06  & 0.54±0.41   & 1.26±0.13&  \textbf{\color{blue}6.63±4.09}  \\
Hammer-C&   1.50 ± 0.69&  \textbf{3.28±3.76 } &   0.67±0.31  &  0.67±0.55  & 0.73±0.21   & 2.14±1.55&  \textbf{4.23±4.40}  \\
\midrule
{Adroit Average}& 28.59 & 20.09 & 21.14 & 21.32 & 22.13 & {\color{blue}25.18}& \textbf{\color{blue}28.97}\\
\bottomrule 
\end{tabular}}
\caption{Offline Preference-based RL in CS-label with IQL algorithm}
\label{result:table_iql_cs}
\end{center}
\end{table*}

\begin{table*}[h]
\begin{center}
\small
\setlength{\tabcolsep}{4pt}{
\begin{tabular}{c|c|cccc|cc}
\toprule  
Dataset& Oracle& MLP& MLP(paper)& TFM& TFM(paper)& DPR& C-DPR\\
\midrule  
HalfCheetah-M&  48.10 ± 0.18&  42.31±0.91&  34.8±5.3& \textbf{46.91±0.37}& \textbf{46.62±6.45} & 44.10±0.2 & 45.09±0.17\\
HalfCheetah-MR& 44.84 ± 0.59&  39.65±0.54&  38.9±0.6& 41.71±0.6& 29.58±5.64 & 40.38±0.53 & \textbf{\color{blue}43.48±0.55}\\
HalfCheetah-ME& 90.78 ± 6.04&  70.69±6.7&  73.8±2.1& \textbf{93.49±1.11}& 80.83±3.46 & \textbf{93.86±0.78} & \textbf{\color{blue}94.58±0.54}\\
\midrule
Hopper-M&  60.37 ± 3.49&  10.00±12.95 &  48.0±40.5& 22.35±11.0& \textbf{99.42±0.94} & 83.54±7.12 & 90.99±4.28\\
Hopper-MR& 64.42 ± 21.52&  46.68±21.45&  25.8±12.3& 87.07±7.6& 41.44±20.88 & \textbf{\color{blue}92.62±2.09} & {\color{blue}90.72±3.81}\\
Hopper-ME& 101.17 ± 9.07&  65.52±24.77&  97.4±15.3& 29.47±26.42&  91.18±27.56 & \textbf{\color{blue}101.95±5.8}& 96.56±10.17\\
\midrule
Walker2D-M&  82.71 ± 4.78&  46.00±21.99&  26.3±13.5& 59.41±32.67& 84.11±1.38 & 82.43±0.88 & \textbf{\color{blue}85.70±0.43}\\
Walker2D-MR& 85.62 ± 4.01&  21.21±17.71&  47.2±25.5& 33.78±20.44& 61.94±11.8 & {\color{blue}67.10±11.93} & \textbf{\color{blue}73.69±1.88}\\
Walker2D-ME& 110.03 ± 0.36&  70.78±45.92&  74.5±59.1& 70.076±41.38& \textbf{110.75±0.9}& 109.77±0.63& \textbf{110.09±0.60}\\
\midrule
{MuJoCo Average}& 76.45&  45.87& 51.85& 51.77& 71.76& \textbf{\color{blue}79.53}& \textbf{\color{blue}81.12}\\
\midrule
Pen-H& -3.88 ± 0.21&  -3.15±0.79&  -3.71±0.20&  -0.99±5.73&  -2.81±2.00 & -3.9±0.32& -3.7±0.06\\
Pen-C& 5.13 ± 5.28&   5.30±7.57&  6.71±7.66 &  3.06±6.92&  \textbf{19.13±12.21}&15.03±11.93& 2.3±5.9\\
\midrule
Door-H&  -0.33 ± 0.01&  -0.26±0.13&  -0.33±0.01&  -0.31±0.08&  -0.32±0.02 & -0.33±0.01& -0.33±0.01\\
Door-C&  -0.34 ± 0.01& -0.33±0.00 &  -0.34±0.00&  -0.33±0.00&  -0.34±0.00 &-0.33±0.00& -0.33±0.00\\
\midrule
Hammer-H&  1.02 ± 0.24&   0.86±0.23&  0.46±0.07 &  0.75±0.40&  1.02±0.41  & 0.89±0.39& 0.87±0.24\\
Hammer-C&  0.25 ± 0.01&   0.34±0.27&  0.45±0.23 &  0.25±0.01&  0.35±0.17  & 0.24±0.00& 0.24±0.00\\
\midrule
{Adroit Average}& 0.24& 0.46& 0.54& 0.40& \textbf{2.83}& 1.93& -0.95\\
\bottomrule 
\end{tabular}}
\caption{Offline Preference-based RL in CS-label with TD3BC algorithm}
\label{result:table_td3bc_cs}
\end{center}
\end{table*}

\begin{table*}[h]
\begin{center}
\small
\setlength{\tabcolsep}{4pt}{
\begin{tabular}{c|c|cccc|cc}
\toprule  
Dataset& Oracle& MLP& MLP(paper)& TFM& TFM(paper)& DPR& C-DPR\\
\midrule  
HalfCheetah-M&   47.04 ± 0.22&          43.7±0.18&   43.9±0.2&   44.14±0.22&  43.26±0.25&   44.88±0.25&  \textbf{\color{blue}45.87±0.28}\\
HalfCheetah-MR&  45.04 ± 0.27&   42.53±0.49&   42.8±0.3&   42.50±0.39& 40.73±2.14&   {\color{blue}43.11±0.27}&  
 \textbf{\color{blue}44.67±0.36}\\
HalfCheetah-ME&  95.63 ± 0.42&   75.30±7.55&   69.0±7.8&   \textbf{93.07±1.6}&  63.84±7.89&   94.7±0.49&  \textbf{94.85±0.93}\\
\midrule
Hopper-M&     59.08 ± 3.77&          57.03±3.65&   57.1±6.5&   57.02±4.24&  44.04±6.92&   \textbf{\color{blue}59.35±3.33}&  \textbf{\color{blue}58.42±2.6}\\
Hopper-MR&    95.11 ± 5.27&   2.25±0.37&    2.1±0.4&    1.81±0.007&   2.08±0.46 &   \textbf{\color{blue}94.71±2.26}&  {\color{blue}94.85±0.93}\\
Hopper-ME&    99.26 ± 10.91&   57.63±4.12&   57.5±2.9&   58.51±4.33&  57.27±2.85&   \textbf{\color{blue}109.23±4.85}& {\color{blue}106.49±4.91}\\
\midrule 
Walker2D-M&   80.75 ± 3.28&          76.01±7.58&   76.9±2.3&   75.53±3.44&  75.62±2.25&   \textbf{\color{blue}83.02±0.7}&  {\color{blue}78.54±3.1}\\
Walker2D-MR&  73.09 ± 13.22&   46.49±24.82&   -0.3±0.0&   61.55±10.06&  33.18±32.35&  \textbf{\color{blue}68.88±5.93}&  \textbf{\color{blue}79.79±5.03}\\
Walker2D-ME&  109.56 ± 0.39&   97.09±14.43&   108.9±0.4&  103.12±13.4&  108.83±0.32& \textbf{109.54±0.37}& \textbf{109.72±0.32}\\
\midrule
MuJoCo Average&  78.28&  55.33& 50.9&  54.26& 52.09& \textbf{\color{blue}78.49}& 
 \textbf{\color{blue}77.88}\\
\midrule
Pen-H&    13.71 ± 16.98&   12.06±20.89&  9.80±14.08&   10.73±14.90&   20.31±17.11& {\color{blue}24.56±6.41} & \textbf{{\color{blue}26.22±6.94}} \\
Pen-C&    1.04 ± 6.62&   1.59±5.00&  \textbf{3.82±8.51} &   3.09±7.55&   \textbf{3.72±7.71}  &  3.34±6.14 & 3.6±6.59\\
\midrule
Door-H&   5.53 ± 1.31&   \textbf{6.37±4.01}&  4.68±5.89 &   5.63±5.06&   4.92±6.22  & 3.42±1.02 & 4.61±0.20 \\
Door-C&    -0.33 ± 0.01&   0.33±0.00&  -0.34±0.00&   0.33±0.00&   -0.34±0.00 & -0.33±0.00 & -0.33±0.00 \\
\midrule
Hammer-H& 0.14 ± 0.11&   0.88±0.32&  0.85±0.27 &   0.86±0.11&   \textbf{1.41±1.14}  &  0.46±0.28 & 0.91±0.14\\
Hammer-C& 0.30 ± 0.01&   0.27±0.01&  0.28±0.01 &   0.28±0.01&   0.29±0.01  &  0.29±0.01 & 0.28±0.00 \\
\midrule
Adroit Average&  3.40&  3.58& 3.18&  3.48& \textbf{5.05}& \textbf{5.28} & \textbf{5.88}\\
\bottomrule 
\end{tabular}}
\caption{Offline Preference-based RL in ST-label with CQL algorithm}
\label{result:table_cql_ST}
\end{center}
\end{table*}

\begin{table*}[h]
\begin{center}
\small
\setlength{\tabcolsep}{4pt}{
\begin{tabular}{c|c|cccc|cc}
\toprule  
Dataset& Oracle& MLP& MLP(paper)& TFM& TFM(paper)& DPR& C-DPR\\
\midrule  
HalfCheetah-M&   48.31 ± 0.22&  45.75±0.17&  47.0±0.1&    46.05±0.05&  45.10±0.17&    47.73±0.18& \textbf{\color{blue}49.05±0.21} \\
HalfCheetah-MR&  44.46 ± 0.22&  42.12±1.42&  \textbf{43.0±0.6}&    41.27±1.13&  40.63±2.42&    \textbf{43.70±0.3} &  \textbf{43.6±1.76}\\
HalfCheetah-ME&  94.74 ± 0.52&  90.81±2.89&  \textbf{92.2±0.5}&    89.34±3.23&  \textbf{92.91±0.28}&    91.338±2.18& 87.93±5.78\\
\midrule
Hopper-M&    67.53 ± 3.78&         50.46±2.61&  51.8±2.8&    50.32±1.48&  37.47±10.57&   \textbf{\color{blue}52.55±3.26}& \textbf{\color{blue}52.92±3.37}\\
Hopper-MR&   97.43 ± 6.39&  56.51±5.92&  70.1±14.3&   52.21±3.04&  64.42±21.52&   \textbf{\color{blue}91.75±1.95}& {\color{blue}72.12±7.48}\\
Hopper-ME&   107.42 ± 7.80&  109.06±3.77& 107.7±3.8&   41.29±24.64&  109.16±5.79&   \textbf{\color{blue}110.28±1.66}& \textbf{\color{blue}110.58±3.75}\\
\midrule 
Walker2D-M&  80.91 ± 3.17&         81.42±2.48&  73.7±11.9&   67.62±6.71&  75.39±6.96&    \textbf{\color{blue}85.41±1.72} &64.46±5.37\\
Walker2D-MR& 82.15 ± 3.03&  73.30±6.23&  68.6±4.3&    59.94±9.53&  60.33±10.31&   \textbf{\color{blue}76.22±6.66}& \textbf{\color{blue}75.22±4.66}\\
Walker2D-ME& 111.72 ± 0.86&  109.94±0.05& 109.8±0.1&   109.41±0.088&  109.16±0.08&  \textbf{\color{blue}112.19±0.19}& \textbf{\color{blue}112.42±0.33}\\
\midrule
MuJoCo Average&  81.63&  73.25&  73.7&  56.005&  69.9&  \textbf{\color{blue}79.01}&74.25 \\
\midrule
Pen-H&    78.49 ± 8.21&    59.63±14.71&  50.15±15.81&  62.75±20.30&  63.66±20.96& \textbf{\color{blue}89.64±12.89} & {\color{blue}86.6±9.00} \\
Pen-C&    83.42 ± 8.19&    71.14±26.51&  59.92±1.12 &  61.77±18.43&  64.65±25.4&  \textbf{\color{blue}74.14±9.46} & {\color{blue}72.43±7.78} \\
\midrule
Door-H&   3.26 ± 1.83&    3.24±1.87&  3.46±3.24  &  4.85±3.51&  \textbf{6.8±1.7}& \textbf{7.25±3.12} & \textbf{6.43±2.00}     \\
Door-C&   3.07 ± 1.75&    -0.03±0.01&  -0.08±0.02 &  -0.1±0.01&  -0.06±0.02& -0.02±0.009 & 0.44±0.68   \\
\midrule
Hammer-H& 1.79 ± 0.80&    0.84±0.42&  \textbf{1.43±1.04}  &  0.67±0.15&  \textbf{1.85±2.04}&  \textbf{1.28±0.09} & 0.83±0.11  \\
Hammer-C& 1.50 ± 0.69&    \textbf{3.48±4.13}&  0.70±0.33  &  0.45±0.26&  1.87±1.5&   2.04±0.14 & 1.19±0.13  \\
\midrule
Adroit Average&  28.59&   23.05&  19.26&  21.74&  23.12& \textbf{\color{blue}29.02} & {\color{blue}27.99} \\
\bottomrule 
\end{tabular}}
\caption{Offline Preference-based RL in ST-label with IQL algorithm}
\label{result:table_iql_ST}
\end{center}
\end{table*}

\begin{table*}[h]
\begin{center}
\small
\setlength{\tabcolsep}{4pt}{
\begin{tabular}{c|c|cccc|cc}
\toprule  
Dataset& Oracle& MLP& MLP(paper)& TFM& TFM(paper)& DPR& C-DPR\\
\midrule  
HalfCheetah-M&  48.10 ± 0.18&          47.51±0.22&  50.3±0.4&   \textbf{51.3±0.419}& 48.06±6.27& 46.25±0.13 &   \textbf{51.04±0.55}\\
HalfCheetah-MR& 44.84 ± 0.59&   44.45±0.5&  44.2±0.4&   43.95±0.87& 36.87±0.48&  41.59±0.51&  \textbf{\color{blue} 45.49±0.36}\\
HalfCheetah-ME& 90.78 ± 6.04&   \textbf{94.28±1.77}&  \textbf{94.1±1.1}&   92.49±2& 78.99±6.82& \textbf{94.47±0.69} &   \textbf{94.53±0.75}\\
\midrule
Hopper-M&      60.37 ± 3.49&          56.46±6.44&  58.6±7.8&   35.8±8.36& 62.89±20.53&  \textbf{\color{blue}65.75±4.92}&   56.46±4.8\\
Hopper-MR&     64.42 ± 21.52&   45.17±8.41&  44.4±18.9&  38.7±1.17& 24.35±7.44 &  \textbf{\color{blue}88.51±9.56} &   {\color{blue}54.79±1.95}\\
Hopper-ME&     101.17 ± 9.07&   84.41±22.55&  103.7±8.3&  \textbf{110.22±2.11}& 104.14±1.82&  107.91±6.04&  82.50±25.88\\
\midrule 
Walker2D-M&     82.71 ± 4.78&          85.03±2.5&  \textbf{86.0±1.5 }&  54.07±24.65& 80.26±0.81 &  84.53±0.3&   84.35±1.46\\
Walker2D-MR&    85.62 ± 4.01&   \textbf{85.93±4.63}&  82.8±15.4&  37.38±33.21& 24.3±8.22  &  65.75±21.36&   82.62±3.07\\
Walker2D-ME&    110.03 ± 0.36&   \textbf{110.32±0.4}&  \textbf{110.4±0.9}& 108.57±4.1& \textbf{110.13±1.87}& 109.60±0.38&  \textbf{110.31±0.5}\\
\midrule
MuJoCo Average&  76.45&  73.11&  74.8& 63.6&  63.33& \textbf{\color{blue}78.27}  &  73.54\\
\midrule
Pen-H&    -3.88 ± 0.21&  -1.34±4.88&  -3.94±0.27 & -0.34±6.68&  -3.94±0.24& -4.04±0.27 & -3.09±1.19\\
Pen-C&    5.13 ± 5.28&  9.86±18.77&  10.84±20.05& 1.80±8.09& \textbf{14.52±16.07}& 6.48±6.79 & 10.75±6.55 \\
\midrule
Door-H&   -0.33 ± 0.01&  -0.32±0.01&  -0.32±0.03 & -0.29±0.04& -0.32±0.03 & -0.34±0.02 & -0.33±0.00\\
Door-C&   -0.34 ± 0.01&  -0.34±0.01&  -0.34±0.00 & -0.33±0.00& -0.34±0.00 & -0.339±0.00 & -0.34±0.00 \\
\midrule
Hammer-H&  1.02 ± 0.24&  0.68±0.27&  1.00±0.12  & 0.78±0.20& 0.96±0.29  & 1.04±0.23 & 1.13±0.11\\
Hammer-C&  0.25 ± 0.01&  0.23±0.01&  0.25±0.01  & 0.24±0.02& 0.27±0.05  & 0.24±0.00 & 0.24±0.00\\
\midrule
Adroit Average&  0.24&  1.46&  1.24&  0.31&  \textbf{1.85}& 0.50 & 1.39\\
\bottomrule 
\end{tabular}}
\caption{Offline Preference-based RL in ST-label with TD3BC algorithm}
\label{result:table_td3bc_ST}
\end{center}
\end{table*}


\begin{table}[htp]
\begin{center}
\setlength{\tabcolsep}{4pt}{
\begin{tabular}{c|cc}
\toprule  
Dataset& DTR&  DTR(C-DPR)\\
\midrule  
HalfCheetah-M&    42.66±0.33& 43.31±0.23\\
HalfCheetah-MR&   40.04±0.54& 40.12±0.29\\
HalfCheetah-ME&   87.47±3.14& \textbf{\color{blue}94.04±0.94}\\
\midrule
Hopper-M&    \textbf{91.75±3.81}& 86.94±6.69\\
Hopper-MR&   \textbf{93.04±1.22}& 80.99±6.18\\
Hopper-ME&   110.85±1.59& 111.04±1.34\\
\midrule
Walker2D-M&    79.01±0.81& \textbf{\color{blue}87.37±0.48}\\
Walker2D-MR&   79.00±3.76& \textbf{\color{blue}86.67±2.75}\\
Walker2D-ME&   109.58±0.52& \textbf{\color{blue}112.22±0.42}\\
\midrule
MuJoCo Average&   81.49&  \textbf{\color{blue}82.52} \\
\bottomrule 
\end{tabular}}
\caption{Offline PbRL in CS-label with DTR algorithm}
\label{appendix:table_DTR_all}
\end{center}
\end{table}

\begin{table*}[htp]
\begin{center}
\small
\setlength{\tabcolsep}{2pt}{
\begin{tabular}{c|cc|cc|cc|cccc}
\toprule  
\multirow{2}*{Dataset}&\multicolumn{2}{|c}{CQL}&  \multicolumn{2}{|c}{IQL}& \multicolumn{2}{|c|}{TD3BC}& \multirow{2}*{FTB}& \multirow{2}*{DPPO}& \multirow{2}*{PT}& \multirow{2}*{IPL}\\
\cmidrule{2-7} 
~ & DPR& C-DPR& DPR& C-DPR& DPR& C-DPR& ~& ~& ~& ~\\
\midrule  
HalfCheetah-MR&   42.67±0.32&	43.62±0.47&	42.29±0.18&	43.88±0.33&	40.35±0.5&	42.87±0.66&	39.0±1.0& 40.8±0.4& \textbf{44.4}& 42.5\\
HalfCheetah-ME&   94.38±0.73&	\textbf{95.49±0.49}& 88.06±4.93&	87.44±3.34&	93.63±0.47& 72.21±13.46&	91.3±1.6& 92.6±0.7& 87.5& 87.0\\
\midrule
Hopper-MR&    75.98±14.04&	87.88±1.09&	84.91±2.46&	86.48±1.12&	75.49±22.86&	88.84±4.8&	\textbf{90.8±2.6}& 73.2±4.7& 84.5& 73.6\\
Hopper-ME&   73.98±6.84&	74.24±1.9&	104.36±5.56&	102.64±5.09&	104.04±2.73&	102.84±4.46&	\textbf{110.0±2.3}& 107.2±5.2& 69&  74.5\\
\midrule
Walker2D-MR&   62.97±6.28&	79.63±4.26&	71.37±3.76&	86.01±1.56&	74.13±4.58&	\textbf{89.45±1.53}&	79.9±5.0& 50.9±5.1& 71.2& 60\\
Walker2D-ME&   109.33±0.50&	109.55±0.37&	110.87±0.63&	\textbf{112.37±0.45}&	109.98±0.54& 110.57±0.82&	109.1±0.1& 108.6±0.1& 110.1& 108.5\\
\midrule
{MuJoCo Average}&   76.55& 81.73& 83.64& \textbf{86.47}& 82.93& 84.46& \textbf{86.68}& 78.88& 77.78& 74.35\\
\bottomrule 
\end{tabular}}
\caption{Comparison results with offline PbRL}
\label{result:table_PbRL_all}
\end{center}
\end{table*}

\begin{table*}[htp]
\begin{center}
\setlength{\tabcolsep}{4pt}{
\begin{tabular}{c|c|cc|cc|c}
\toprule  
Dataset& MLP& DPR& C-DPR&  Oracle\\
\midrule  
HalfCheetah-M&   -257.66±287.73&	-215.47±7.31&	\textbf{-126.32±137.08}&	    \multirow{3}*{1818.95±1011.79}\\
HalfCheetah-MR&  -1024.83±699.36&	941.68±1242.71&	\textbf{2516.80±1622.77}$^{\dagger}$&	~\\
HalfCheetah-ME&  -2569.83±1311.26&	\textbf{-183.66±14.87}&	-669.92±349.75&	    ~\\
\midrule
Hopper-M&    966.42±71.69&	 \textbf{1665.65±1215.15}&   1137.32±293.27&	 \multirow{3}*{1316.42±209.80}\\
Hopper-MR&   240.92±177.86&	 1620.77±856.31&	\textbf{2297.24±628.75}&	 ~\\
Hopper-ME&   887.93±155.64&  \textbf{2724.95±956.76}$^{\dagger}$&	2238.53±456.11&	 ~\\
\midrule
Walker2D-M&    93.86±155.34&	1029.90±28.65&	\textbf{1363.72±384.32}&	\multirow{3}*{1927.02±450.30$^{\dagger}$}\\
Walker2D-MR&   351.62±147.03&	\textbf{1016.14±27.12}&	989.47±151.65&	~\\
Walker2D-ME&   -5.97±0.46&	    986.82±27.01&	\textbf{1037.91±230.22}&	~\\
\midrule
{MuJoCo Average}&   -146.39& 1065.19& 821.55& 1687.46$^{\dagger}$\\
\bottomrule 
\end{tabular}}
\caption{Comparison results with PPO in online scenarios. "Oracle" denotes the environment's built-in rewards. We use \textbf{bold} to indicate the best results in single dataset and use $^{\dagger}$ to indicate the best results in single online scenario.}
\label{result:table_ppo_all}
\end{center}
\end{table*}

\begin{table}[htp]
\begin{center}
\setlength{\tabcolsep}{4pt}{
\begin{tabular}{c|ccc|ccc}
\toprule  
Dataset& DPR(2)& (50)& (2000)& C-DPR(2)& (50)& (2000)\\
\midrule  
HalfCheetah-M&    42.72±0.43&	\textbf{44.37±0.11}&	\textbf{44.90±0.12}&	42.69±0.42&	 \textbf{47.34±0.35}&	45.57±0.13\\
HalfCheetah-MR&   36.64±2.5&	41.02±0.31&	\textbf{42.79±0.32}&	37.22±1.29&	42.93±0.2&	\textbf{44.09±0.29}\\
HalfCheetah-ME&   60.57±6.11&	\textbf{91.13±0.83}& 87.99±4.66&	58.85±5.69&	91.03±2.59&   \textbf{93.62±0.95}\\
\midrule
Hopper-M&        72.53±1.71&	70.14±1.75&	\textbf{74.49±2.1} &	74.75±2.5&	\textbf{76.99±4.75}&	70.98±3.93	\\
Hopper-MR&       33.87±3.42&	78.25±7.2&	\textbf{87.53±1.06}&	32.8±4.03&	\textbf{89.57±0.66}&	86.10±1.47	\\
Hopper-ME&       96.48±19.57&	104.04±6.73&	\textbf{108.44±6.08}&	88.07±20.34&	\textbf{107.96±5.37}&	99.34±10.07\\
\midrule
Walker2D-M&      75.24±1.67&	78.79±5.95&	\textbf{83.47±1.59}&	76.64±1.23&	78.15±10.53&	\textbf{85.77±0.4}\\
Walker2D-MR&     56.18±12.6&	\textbf{72.38±2.52}&	67.23±8.51 &	64.63±4.73&	\textbf{85.20±1.49}&	75.95±1.72\\
Walker2D-ME&     109.12±0.46&	\textbf{111.89±0.29}&	\textbf{111.89±0.18}&	109.76±0.2&	\textbf{112.36±0.15}&   \textbf{111.98±0.26}\\
\midrule 
MuJoCo Average&  64.81& 76.89&   \textbf{78.75}&  65.04&  \textbf{81.28}& 79.27 \\
\bottomrule 
\end{tabular}}
\caption{Ablation experiments with different batch sizes of preference data. We use \textbf{bold} to indicate the best results of DPR and C-DPR, respectively.}
\label{result:table_batchsize_all}
\end{center}
\end{table}

\begin{table}[t]
\begin{center}
\setlength{\tabcolsep}{7pt}{
\begin{tabular}{c|cc}
\toprule  
Dataset& DPR+IQL& DPR(noiseless)+IQL\\
\midrule  
HalfCheetah-M&    \textbf{44.90±0.12}& 43.07±0.23\\
HalfCheetah-MR&   \textbf{42.79±0.32}& 39.96±0.95\\
HalfCheetah-ME&   87.99±4.66& \textbf{89.72±1.43}\\
\midrule
Hopper-M&    \textbf{74.49±2.1}& 66.98±2.40\\
Hopper-MR&   \textbf{87.53±1.06}& 48.99±12.67\\
Hopper-ME&   \textbf{108.44±6.08}& 104.19±8.12\\
\midrule
Walker2D-M&    \textbf{83.47±1.59}& 79.11±1.36\\
Walker2D-MR&   \textbf{67.23±8.51}& 50.41±15.52\\
Walker2D-ME&   111.89±0.18& 110.91±0.14\\
\midrule
MuJoCo Average&   \textbf{78.75}&  70.37\\
\bottomrule
\end{tabular}}
\caption{Ablation experiments on noise disturbance.}
\label{result:table_noiseless_all}
\end{center}
\end{table}

\clearpage

\subsection{B.2 Training Curve of Offline PbRL}
\label{appendix:result_curve_all}

\clearpage

\subsection{B.3 Timestep of Diffusion Reward Model}
\label{appendix:result_timestep_all}

\begin{figure*}[h]
    \centering
    \subfigure[IQL(DPR)]{
    \includegraphics[width=0.75\linewidth]{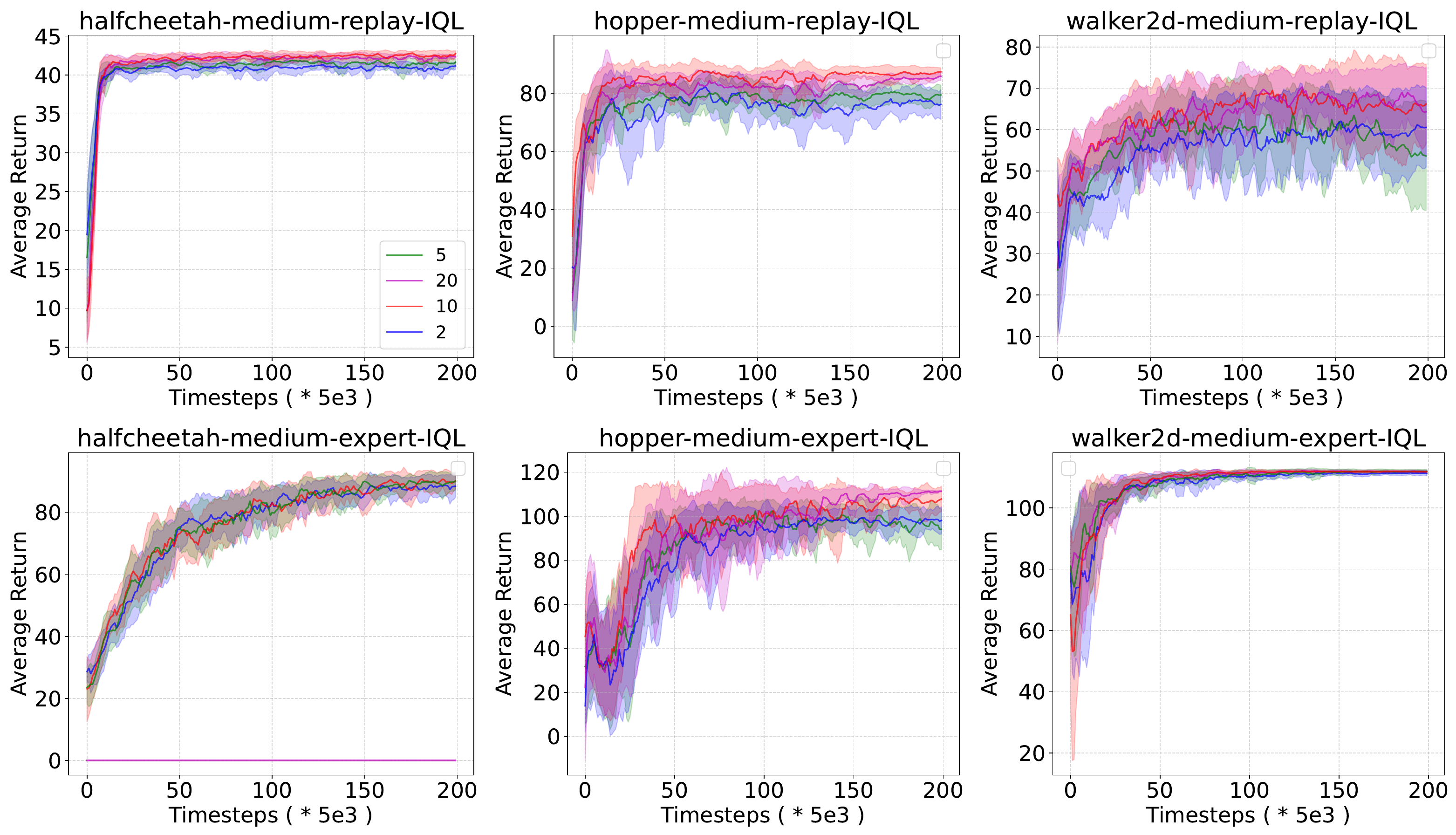}}
    \subfigure[IQL(C-DPR)]{
    \includegraphics[width=0.75\linewidth]{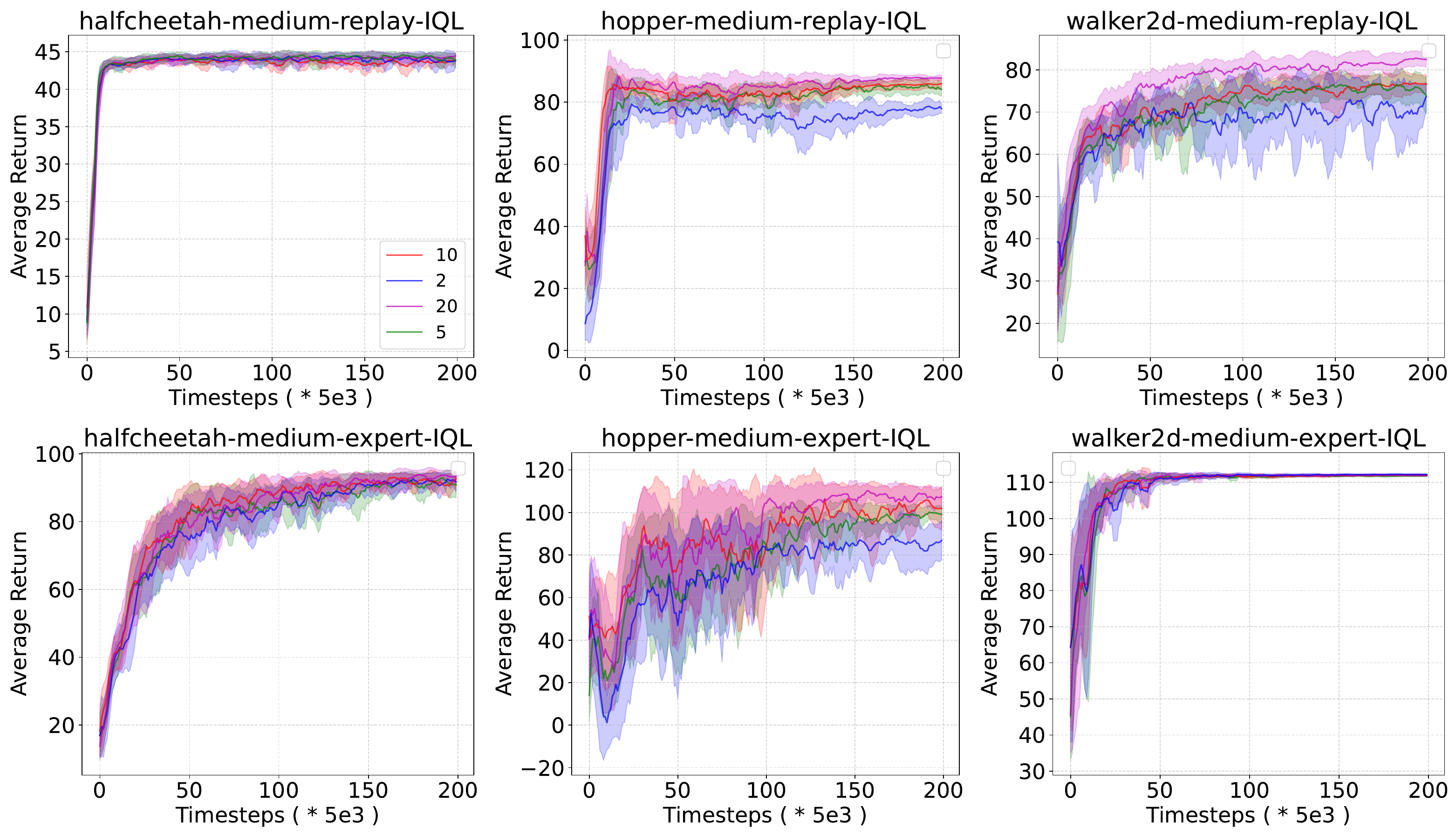}}
    \caption{Training curve of IQL algorithm based on CS-label with different diffusion timestep}
    \label{fig:diffusion_iql}
\end{figure*}

\begin{figure*}[h]
    \centering
    \subfigure[TD3BC(DPR)]{
    \includegraphics[width=0.75\linewidth]{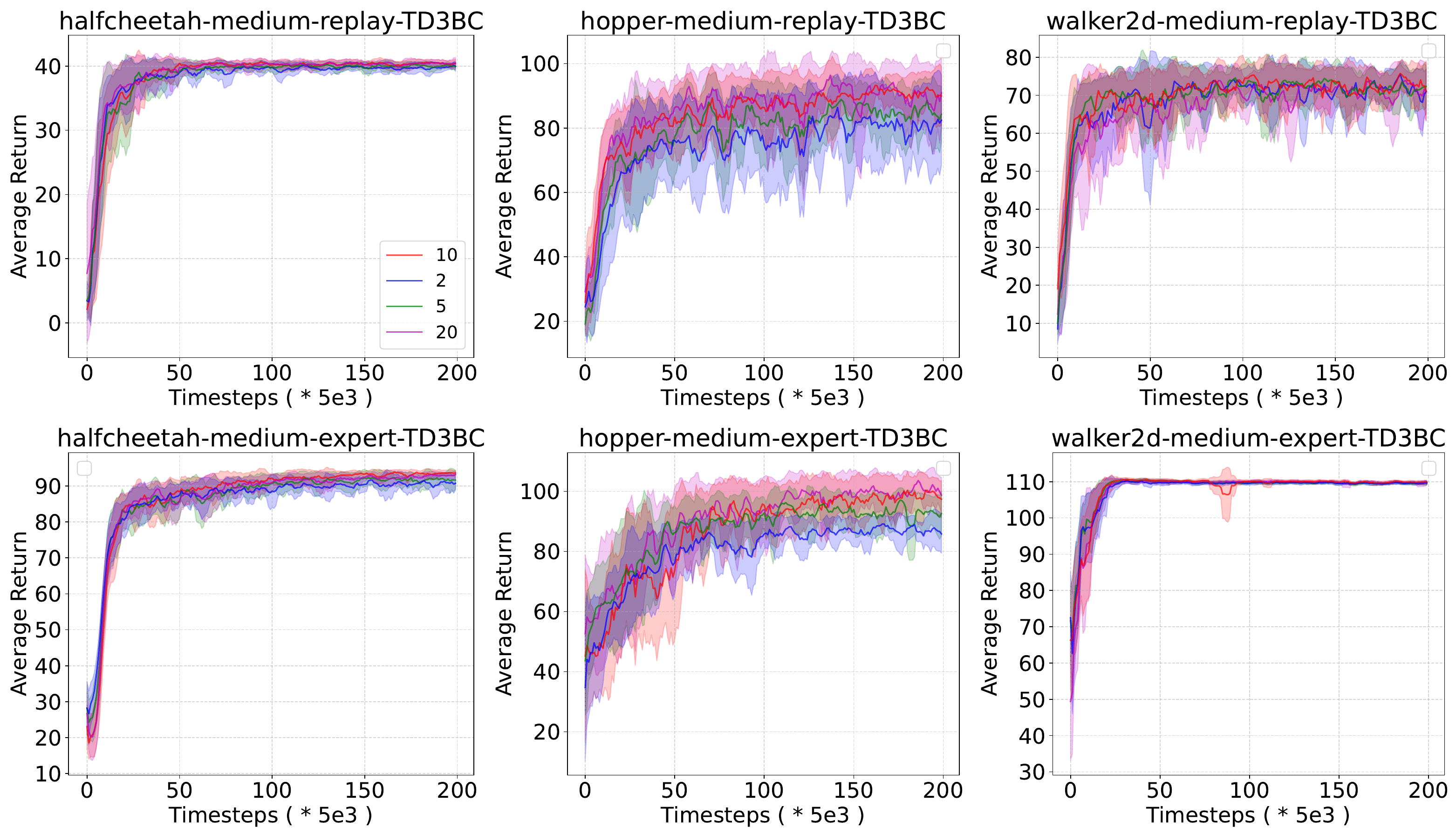}}
    \subfigure[TD3BC(C-DPR)]{
    \includegraphics[width=0.75\linewidth]{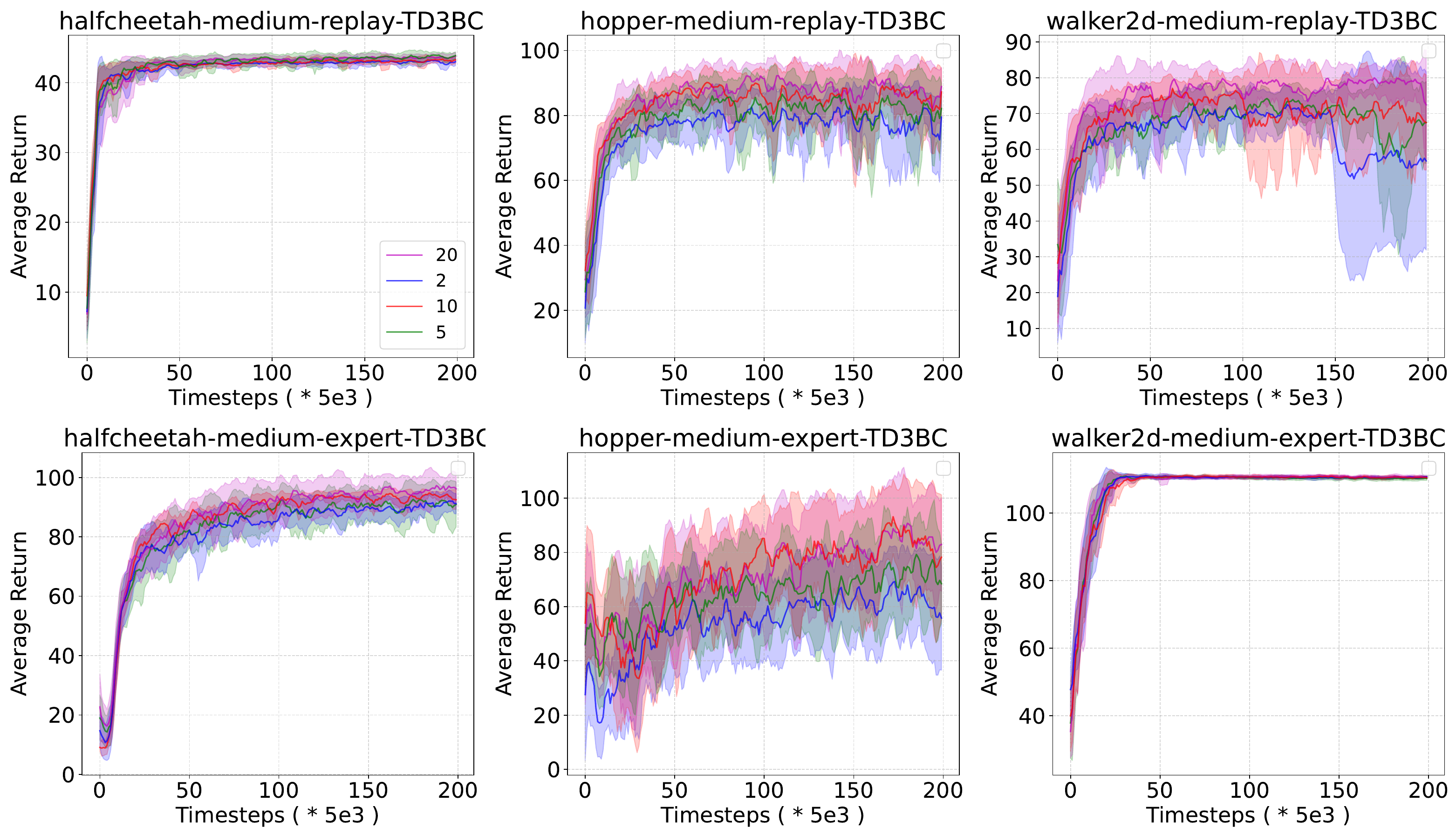}}
    \caption{Training curve of TD3BC algorithm based on CS-label with different diffusion timestep}
    \label{fig:diffusion_td3bc}
\end{figure*}


\end{document}